\documentclass[lettersize,journal]{IEEEtran}
\usepackage{amsmath}
\usepackage{amssymb}

\usepackage{amsthm}
\newtheorem{definition}{Definition}
\usepackage{algorithmicx}
\usepackage{algcompatible}
\usepackage{algorithm}
\usepackage{algpseudocode}
\usepackage{array}
\usepackage[font=normalsize,labelfont=sf,textfont=sf]{subcaption}
\usepackage{textcomp}
\usepackage{hyperref}
\newenvironment{problem}[1][]
  {\par\noindent\textbf{Problem.} \textit{#1} \par\noindent}{\par}
\usepackage{stfloats}
\usepackage{url}
\usepackage{verbatim}
\usepackage{graphicx}
\usepackage{subcaption}
\usepackage[utf8]{inputenc}
\usepackage{newunicodechar}
\newunicodechar{∗}{\ast}
\usepackage{cite}
\usepackage{booktabs}
\usepackage{multirow}
\usepackage{adjustbox}
\usepackage{tikz}
\usepackage{placeins}  
\usepackage{graphicx}   
\usepackage{pdfpages}

\usepackage[utf8]{inputenc}
\begin{document}

\title{DARTS-GT: Differentiable Architecture Search for Graph Transformers with Quantifiable Instance-Specific Interpretability Analysis}

\author{Shruti Sarika Chakraborty
\thanks{Department of Computer Science, University of Oxford, Oxford OX1 3QG, UK. Emails: \texttt{shruti.chakraborty@cs.ox.ac.uk}, \texttt{peter.minary@cs.ox.ac.uk}.}
        \and
        Peter Minary}

\markboth{Journal of \LaTeX\ Class Files,~Vol., No., October~2025}%
{Shell \MakeLowercase{\textit{et al.}}: DARTS-GT: Differentiable Architecture Search for Graph Transformers}

\maketitle

\begin{abstract}
Graph Transformers (GTs) have emerged as powerful architectures for graph-structured data, yet remain constrained by rigid designs and lack quantifiable interpretability methods. Current state-of-the-art GTs commit to fixed GNN types across all layers, missing potential benefits of depth-specific component selection, while their increasingly complex architectures become opaque black boxes where performance gains cannot be distinguished between meaningful structural patterns and spurious correlations. We redesign the GT attention mechanism through asymmetry, decoupling structural encoding from feature representation. Queries derive directly from node features, while keys and values come from graph neural network (GNN) transformations, separating how the model learns features from how it encodes graph structure. Within this asymmetric framework, we use Differentiable ARchiTecture Search (DARTS) to select optimal GNN operators at each layer, enabling depth-wise heterogeneity inside the transformer attention itself, hence the name DARTS-GT.
To understand these discovered architectures, we develop the first quantitative interpretability framework for GTs through causal ablation that identifies which heads and nodes actually drive predictions. Our metrics: Head-deviation, Specialization, and Focus, reveal the specific components responsible for each prediction while enabling broader model comparison.  Experiments across eight benchmarks demonstrate that DARTS-GT achieves state-of-the-art performance on four datasets while remaining competitive on others, with discovered architectures revealing dataset-specific patterns ranging from highly specialized to balanced GNN distributions. Our interpretability analysis reveals that visual attention salience and causal importance do not necessarily correlate, indicating that widely used visualization approaches may miss the components that actually matter for predictions. Crucially, the heterogeneous architectures found by DARTS-GT consistently produced more interpretable models than baseline GTs, establishing that Graph Transformers do not need to choose between performance and interpretability.
\end{abstract}


\section{Introduction}

For graph-structured data, Graph Transformers (GTs) have become a dominant architectural choice, combining attention mechanisms with graph-awareness \cite{lin_mesh_2021, chen_structure-aware_2022}. Their success spans protein structure-to-function prediction~\cite{gu_hierarchical_2023}, drug discovery~\cite{gao_graphormerdti_2024}, and materials design~\cite{yuan_survey_2025}, where understanding complex structural patterns is crucial.

Current state-of-the-art GTs incorporate graph structure through GNN-Transformer combinations~\cite{rampa_recipe_nodate,chen_structure-aware_2022}, specialized positional encodings~\cite{kreuzer_rethinking_2021}, and attention augmentation with structural biases~\cite{ying_transformers_2021}.  However, these architectures commit to a fixed GNN type throughout all layers or a predetermined encoding scheme. This rigidity prompts an investigation into whether GTs benefit from depth-specific component selection.

Depth-specific, task-adaptive GNN selection is motivated by recent neural-architecture-search (NAS) successes in GNNs. Methods like GraphNAS~\cite{gao_graphnas_2019}, SANE~\cite{zhao_search_2021}, and PAS~\cite{wei_neural_2024} show that heterogeneous architectures employing different GNN operations across depths outperform homogeneous models. These works reveal that optimal architectures develop dataset-adaptive depth-wise specialization, learning complementary patterns at different scales. In GTs, where multiple attention heads already capture diverse node-node interactions, depth-wise architectural diversity becomes even more compelling, yet remains unexplored.

Relaxing architectural constraints introduces a broader challenge. As Graph Transformer variants become more flexible, their internal decision processes become increasingly opaque. Without interpretability methods, it is difficult to determine whether reported gains reflect meaningful structural patterns or spurious correlations. This, for example, forms a crucial distinction in scientific domains such as drug discovery, where predictions must be justifiable. However, this issue is not specific to a particular GT design: for instance, in GraphGPS (also referred to as GPS)~\cite{rampa_recipe_nodate}, which has a hybrid-GT architecture by combining message-passing-neural-networks (MPNN) in parallel with transformer attention,  it is unclear whether improvements arise from the auxiliary MPNN or the Transformer path. Similarly, in Graph-Trans ~\cite{wu_representing_2022} and SAT ~\cite{chen_structure-aware_2022}, the contribution of the GNN pre-encoding or k-subtree extraction, respectively, is similarly entangled. In short, regardless of the GT approach, interpretability is essential to clarify which components and structures truly drive model performance.


Recent NAS efforts for GTs include EGTAS  ~\cite{wang_automatic_2024}, AutoGT  ~\cite{zhang_autogt_2023}, UNIFIEDGT  ~\cite{lin_unifiedgt_2024} and UGAS  ~\cite{song_towards_2025}. EGTAS uses evolutionary search to explore topologies and graph-aware strategies by searching through established architectural paradigms from state-of-the-art designs. This essentially navigates between existing construction patterns,  selecting in this process a single GNN block type, applied uniformly across depth. AutoGT employs evolutionary strategies over established GT templates and encodings, without depth-wise GNN selection. UNIFIEDGT exposes modular search over components and selects an overall configuration rather than layer-wise GNN selection within attention blocks. UGAS uses Differentiable ARchiTecture Search (DARTS) or Genetic Algorithms (GA) to search over subsets of MPNN modules (e.g., GCN~\cite{kipf_semi-supervised_2017}, GAT~\cite{velickovic_graph_2018}) and GT modules (Transformer~\cite{dwivedi_generalization_2021}, Performer~\cite{choromanski_rethinking_2022}) deployed in parallel within each layer. Each layer may contain only MPNNs, only GTs, or any mixture, with GTs treated as optional candidates rather than architectural foundations.

The selection of depth-specific GNN presents a formidable optimization problem: with $n$ GNN variants across $L$ layers, the search space contains $n^L$ configurations. Thus, systematic exploration of this search space is infeasible in most practical applications. An effective search must intelligently adapt to datasets and tasks, determining which GNN types to deploy and their optimal layer positions (depth). This requires automated search techniques, specifically NAS, capable of navigating this vast configuration space.

These discovered architectures must also be interpretable for trustworthiness and scientific validity. Unlike well-developed GNN explainability methods~\cite{ying_gnnexplainer_2019} that identify prediction-critical subgraphs, GT interpretation, widely regarded as a blackbox~\cite{shehzad_graph_2024}, remains dominated by attention visualization~\cite{hussain_global_2022} lacking quantitative metrics and the ability to differentiate visually prominent but functionally irrelevant patterns from subtle yet critical ones. More recently, noteworthy efforts in GT interpretability have focused on analyzing attention graphs (i.e. aggregating head/layer attention into an attention graph)~\cite{el_towards_2025}.
However, none of the prior methods test faithfulness (whether heads/nodes causally affect predictions) nor enable cross-model comparison via standardized scores. Task-specific self-explanations exist but do not generalize to arbitrary GTs or provide head attributions~\cite{li_self-explainable_2024}. We seek a GT-agnostic instance-level interpretability framework grounded in perturbation/ablation that delivers model-level scores for comparing GT interpretability.

Hence, our investigation addresses two complementary research questions:
\textit{Q1: Can Graph Transformers achieve improved performance and adaptability by dynamically selecting GNN types per layer (depthwise) and creating dataset-specific architectures ranging from uniform to heterogeneous?}
\textit{Q2: How can we quantitatively assess the interpretability of a GT and identify which regions of the graph and architectural components drive specific predictions?}

These questions are  inter-linked: Q1 explores performance potential, while Q2 ensures interpretability and trustworthiness, essential for high-stakes domains.

To address these challenges, we introduce a comprehensive framework with two synergistic components:

First, we employ DARTS to select GNN operators within the attention blocks at each layer. In doing so, we reimagine the asymmetric attention mechanism initially proposed by GMTPool~\cite{baek_accurate_2021}, but in a more computationally efficient form: queries are derived directly from node features, while keys and values are obtained through GNN transformations of those features. This design explicitly separates structural encoding from feature representation and enables depth-wise heterogeneity across layers, embedding it directly inside the attention mechanism rather than through external augmentation, thereby rewiring how GTs capture graph structure. The search procedure is powered by DARTS~\cite{liu_darts_2019}, which leverages gradient-based optimization of the continuously relaxed architecture space, avoiding the expense of training discrete candidates from scratch. DARTS has been extensively validated in prior graph NAS studies (e.g., SANE~\cite{zhao_search_2021}, GASSO~\cite{qin_graph_2021}, PAS~\cite{wei_neural_2024}, UGAS), underscoring its computational efficiency and robustness in navigating combinatorial design spaces.

Second, we introduce a Fidelity like~\cite{lukyanov_robustness_2025,amara_graphframex_2024}, instance-level, ablation-grounded score: \textbf{Head-deviation}, quantifying prediction changes when specific heads are masked. From per-head effects on each instance, we compute two complementary metrics: \textbf{Specialization} captures \textit{head distinguishability} as dispersion of head impacts (high = few heads carry causal load; low = heads act interchangeably); \textbf{Focus} captures \textit{consensus} over graph regions by measuring top-head agreement on top-k nodes (high = heads converge on same salient substructures or nodes). Aggregating per-instance values by dataset median yields robust model-level summaries. Together, these form our interpretability framework that we propose to analyse our model along with existing models for comparison. Head-deviation provides head and graph-region attributions per input, while median Specialization and median Focus enable dataset-level GT comparison indicating which model is easier to interpret at matched accuracy.

This interpretation is particularly valuable in applications like drug toxicity prediction. When identifying toxic molecules, practitioners must verify the focus on actual toxicophores versus spurious correlations critical for regulatory approval and scientific validity. Also, without methods to quantify the difficulty of interpretation, meaningful architecture comparison remains impossible. Our framework provides quantifiable metrics representing the first quantitative framework for both globally evaluating and comparing GT interpretability while providing instance-specific analysis that traces predictions from architectural components to graph regions.

Our specific contributions:
\begin{enumerate} 
    \item The first DARTS-based architecture search for depth-wise GNN selection within GT layers, exploring heterogeneity through layer-specific GNN operators within attention blocks. Our approach maintains computational efficiency to find depth-wise optimal GNN selections among the possible configurations ($n$ GNN variants, $L$ layers).  This enables automated  discovery of task optimal architectures.
    
    \item Strong empirical performance achieving competitive or SOTA results on multiple benchmarks. Our architecture consistently outperforms standard GTs~\cite{dwivedi_generalization_2021} across dense and sparse attention mechanisms, while matching or exceeding established baselines: Recent GNNs (e.g. GCN~\cite{kipf_semi-supervised_2017}, GIN~\cite{xu_how_2019}, GatedGCN~\cite{bresson_residual_2018}), recent GT architectures (e.g. GPS~\cite{rampa_recipe_nodate}, Graphormer~\cite{ying_transformers_2021}), and NAS-based approaches (e.g. EGTAS~\cite{wang_automatic_2024}, AutoGT~\cite{zhang_autogt_2023}, UGAS~\cite{song_towards_2025}).
    
    \item To our knowledge,  we also introduce the first GT-agnostic interpretability framework that is both instance-level and quantitative. We compute an ablation-grounded Head-deviation metric per input, then derive Specialization (head distinguishability) and Focus (consensus of top-head node sets). Dataset-level medians provide robust model-level summaries, enabling both local evaluation (which heads/nodes mattered) and global evaluation (which model is easier to interpret). Our analyses reveal that visually broad attention can emerge as functionally irrelevant, while sparse patterns can be decisively predictive: underscoring the limits of visualization-based approaches.
    
    \item We apply the interpretability framework across three architectures (DARTS-GT, GPS~\cite{rampa_recipe_nodate} and standard (Vanilla) GT~\cite{dwivedi_generalization_2021}) and show that the DARTS-GT derived model is more consistent overall in terms of overall consensus (Focus) and head-distinguishability (Specialization).
\end{enumerate}

In summary, we (i) propose a differentiable NAS framework that builds depth-wise heterogeneous Graph Transformers through direct redesigning of the attention mechanism rather than auxiliary GNN augmentation and (ii) develop quantitative interpretability metrics at head- and node-level per instance. 
Our code is available at: \href{https://github.com/shrutiOx/DARTS_GT}{https://github.com/shrutiOx/DARTS\_GT}.

\section{Related Work}

We discuss three interconnected themes related to this work in separate paragraphs.

\textbf{Graph Transformer :} A broad line of recent work combines global self-attention with graph inductive bias (GNNs)~\cite{wu_representing_2022}. GraphGPS~\cite{rampa_recipe_nodate} interleaves local message passing with global attention and aggregates rich positional/structural encodings, effectively in a parallel MPNN and transformer attention design, setting a strong template for hybrid GTs. SAN~\cite{kreuzer_rethinking_2021} injects spectral positional information and runs a fully connected Transformer, showing that learned spectral PEs can offset oversquashing while preserving structure. SAT~\cite{chen_structure-aware_2022} goes further and makes attention itself structure-aware by extracting k-subtree/k-subgraph features with a base GNN before computing attention, consistently improving performance over baseline GTs. Graphormer~\cite{ying_transformers_2021} augments attention with centrality, shortest-path and edge encodings, and demonstrated state-of-the-art molecular property prediction at scale. Variants in other modalities (e.g. Mesh Graphormer~\cite{lin_mesh_2021}) reinforce the pattern of mixing graph convolution with attention to capture local+global interactions. Graph-Trans~\cite{wu_representing_2022} integrates a preliminary stack of GNN layers before Transformer blocks, using the GNN to encode local structural patterns that the subsequent attention layers can then exploit. 
Our work sits in this stream but differs in how the local component is chosen: instead of a fixed GNN extractor or a single hybrid template, we dynamically select the GNN type per Transformer layer to yield dataset-specific, potentially heterogeneous stacks. 

A related pooling-centric line is Graph Multiset Transformer (GMT)~\cite{baek_accurate_2021}, which builds graph-level representations with attention-based pooling. GMT computes K/V via GNNs to explicitly encode structure while forming queries from node embeddings, improving over linear projections used in standard multi-head-attention (MHA). Yet GMT's implementation suffers from computational inefficiency: each attention head requires independent GNN operations for its keys and values, scaling complexity linearly with head count. Moreover, their framework remains confined to graph-level pooling rather than general transformation layers. We borrow the idea of asymmetric query–key/value sources, but reimagine it within full GT layers with a computationally efficient approach: queries come directly from node features (instead of node embeddings in GMT that undergo another layer of GNN), while keys/values are produced by layer-specific GNN operators, shared across heads for efficiency and generalization beyond graph-level pooling.

\textbf{NAS for Graphs and Graph Transformers:} Early Graph neural-architecture-search (GNAS) studies such as GASSO~\cite{qin_graph_2021}, SANE~\cite{zhao_search_2021}, GraphNAS~\cite{gao_graphnas_2019} and PAS~\cite{wei_neural_2024} are searched over message-passing cells on pure GNN based backbone supernets. However, for GTs, AutoGT~\cite{zhang_autogt_2023} defines a unified GT formulation and searches Transformer hyperparameters plus structural/positional encodings, but the resulting architecture is globally tied rather than layer-wise heterogeneous. EGTAS~\cite{wang_automatic_2024} expands the GT search space with macro- (topology) and micro- (graph-aware strategies) design and uses evolutionary search to assemble end-to-end GTs. UGAS~\cite{song_towards_2025} proposes a Unified Graph Neural Architecture Search that explores which subsets of MPNN modules and GT modules to activate in parallel per layer, with GTs treated as optional candidates rather than foundational components. Complementary frameworks such as UNIFIEDGT~\cite{lin_unifiedgt_2024} modularize GT components (sampling, structural priors, attention, local/global mixing, FFNs) to instantiate many GT variants. Our work is orthogonal. We keep the GT spine and search for the local operator (GNN) per layer with DARTS, yielding adaptive uniform/heterogeneous architectures while preserving global attention. This enables direct analysis of how structural encoding choices impact performance within a consistent computational framework.

\textbf{Interpretability of (Graph) Transformers}
Existing transformer studies investigate head importance for pruning~\cite{michel_are_2019}, i.e., measuring global accuracy drops to compress models, or through head ablation to assess model performance under masking~\cite{pochinkov_investigating_nodate},\cite{bahador_mechanistic_nodate}. We instead quantify instance-specific importance for interpretability, identifying which heads and nodes drive individual predictions rather than finding globally redundant components. 
This connects to the broader line of causal interventions known as activation patching, where internal components are ablated or replaced, and the output change is measured~\cite{heimersheim_how_2024}\cite{zhang_towards_2024}. Zero-ablation, as we employ, is a popular variant~\cite{heimersheim_how_2024} amongst other causal-tracing techniques. Our contribution adapts this causal, instance-level paradigm to Graph Transformers, introducing head-specific ablations with novel metrics.
Although GNN explainability has matured through tools like GNNExplainer~\cite{ying_gnnexplainer_2019} that identify prediction-critical subgraphs, graph transformers lack equivalent methodologies. Most GT interpretability either visualizes attention~\cite{hussain_global_2022} or aggregates it across layers. Methods based on attention flow track information paths through attention, but remain correlational~\cite{el_towards_2025}. Lastly, some protein structure to function prediction papers~\cite{gu_hierarchical_2023 } use GradCAM in GT attention to identify 'important' residues, but this only shows where gradients flow, not whether those regions actually
help correct prediction i.e. if it truly contributes to accuracy. Our study complements this space with causal ablations at two levels: (i) Specialization (indicates whether heads are easily distinguishable
by their causal impact) and (ii) Focus (are top-performing head agree on node consensus). The proposed metrics align with established interpretability principles for GNNs (e.g. sparsity, consistency, fidelity)~\cite{lukyanov_robustness_2025} while tying interpretations directly to measured performance changes under controlled deletions.

\section{Methods}

\subsection{Preliminaries}

We consider a graph $G = (V, E)$ with node features $\mathbf{F} \in \mathbb{R}^{n \times d}$, where $n = |V|$ and $d$ is the feature dimension. A Graph Transformer consists of $L$ layers, each containing multi-head attention followed by feed-forward modules. Let $\mathbf{Z}^{(\ell)} \in \mathbb{R}^{n \times d}$ denote the node embeddings in layer $\ell$.

For brevity, we omit the layer index $\ell$ when the context is clear. With $M$ attention heads where $m \in [1,2,...,M]$, the attention components are computed as:
\begin{equation}
\mathbf{Q}_m = \mathbf{Z} \mathbf{W}_{Q,m}, \quad
\mathbf{K}_m = \mathbf{Z} \mathbf{W}_{K,m}, \quad
\mathbf{V}_m = \mathbf{Z} \mathbf{W}_{V,m}
\end{equation}
where $\mathbf{Q}_m, \mathbf{K}_m, \mathbf{V}_m \in \mathbb{R}^{n \times d_m}$, $\mathbf{W}_{Q,m}, \mathbf{W}_{K,m}, \mathbf{W}_{V,m} \in \mathbb{R}^{d \times d_m}$ and $d_m = d/M$. 
We assume that $d_m \in \mathbb{Z}_{>0}$ (i.e. $d$ is divisible by $M$).\\

\subsubsection{Attention Mechanisms} In our study,
we explore dense and sparse attention.

\paragraph{Full Attention:} Each node attends to all others:
\begin{equation}
\mathbf{S}_m = \text{softmax}\left(\frac{\mathbf{Q}_m \mathbf{K}_m^{\top}}{\sqrt{d_m}}\right), \quad
\mathbf{Y}_m = \mathbf{S}_m \mathbf{V}_m
\end{equation}
where $\mathbf{S}_m \in \mathbb{R}^{n \times n}, \mathbf{Y}_m \in \mathbb{R}^{n \times d_m}$ and \texttt{softmax} is applied row-wise for full attention (per-query normalization).  This leads to a complexity of $\mathcal{O}(n^2 d_m)$ per head and $\mathcal{O}(n^2 \sum_m d_m) = \mathcal{O}(n^2 d)$ across $M$ heads.

\paragraph{Sparse Attention} In this formulation, nodes attend only to graph neighbors. This leads to a complexity of $\mathcal{O}(|E| d_m)$ per head and $\mathcal{O}(|E| \sum_m d_m) = \mathcal{O}(|E| d)$ across $M$ heads. For the edge set $E = \{(i, j)\}$ where $i$ and $j$ denote source and target nodes. For each edge $(i, j)$, we compute the attention coefficient between source node $i$ and target node $j$. Thus for edge $(i,j) \in E$:
\begin{equation}
\beta_{m,ij} = \frac{\langle [\mathbf{K}_m]_i, [\mathbf{Q}_m]_j \rangle}{\sqrt{d_m}}
\end{equation}

\begin{equation}
s_{m,ij} = \frac{\exp(\beta_{m,ij})}{\sum_{i': (i',j) \in E} \exp(\beta_{m,i'j})}
\end{equation}

\begin{equation}
[\mathbf{Y}_m]_j = \sum_{i: (i,j) \in E} s_{m,ij} [\mathbf{V}_m]_i
\end{equation}
where $[\mathbf{Y}_m]_j$, $[\mathbf{Q}_m]_j$, $[\mathbf{K}_m]_i$, and $[\mathbf{V}_m]_i$ denote the $j$-th and $i$-th rows of the respective matrices.

\subsubsection{Multi-Head Concatenation}
Each head outputs $\mathbf{Y}_m\in\mathbb{R}^{n\times d_m}$, with $d_m = d/M$.
We concatenate the head outputs along the feature dimension and project back to the model dimension $d$:
\begin{equation}
\label{eq:mh-concat}
\begin{aligned}
\mathbf{Y} &\;=\; [\mathbf{Y}_1 \,\|\, \mathbf{Y}_2 \,\|\, \cdots \,\|\, \mathbf{Y}_M] \;\mathbf{W}_{\text{out}},\\
&\quad\text{where } \mathbf{W}_{\text{out}} \in \mathbb{R}^{(M d_m)\times d} = \mathbb{R}^{d\times d}.
\end{aligned}
\end{equation}

\subsubsection{Feed-Forward Layers and Skip Connections}
Following the attention module, each block applies a two-layer position-wise feed-forward network with residual connections and normalization applied after each sublayer:
\begin{align}
\hat{\mathbf{Z}}^{(\ell)} &= \mathrm{Norm}\!\big(\mathbf{Z}^{(\ell)} + \mathbf{Y}^{(\ell)}\big),\\
\tilde{\mathbf{Z}}^{(\ell)} &= \sigma\!\big(\hat{\mathbf{Z}}^{(\ell)} \mathbf{W}_1^{(\ell)} + \mathbf{b}_1^{(\ell)}\big)\,\mathbf{W}_2^{(\ell)} + \mathbf{b}_2^{(\ell)},\\
\mathbf{Z}^{(\ell+1)} &= \mathrm{Norm}\!\big(\hat{\mathbf{Z}}^{(\ell)} + \tilde{\mathbf{Z}}^{(\ell)}\big),
\end{align}
where $\sigma$ is the nonlinearity (we use \textsc{ReLU} in our implementation), $\mathbf{W}_1^{(\ell)} \in \mathbb{R}^{d \times r d}$, $\mathbf{W}_2^{(\ell)} \in \mathbb{R}^{r d \times d}$, and $\mathbf{b}_1^{(\ell)}, \mathbf{b}_2^{(\ell)}$ are biases. We use $r{=}2$ in our implementation, and $\mathrm{Norm}$ denotes either LayerNorm or BatchNorm depending  on configuration.

\begin{figure}
    \centering
    \includegraphics[width=1\linewidth]{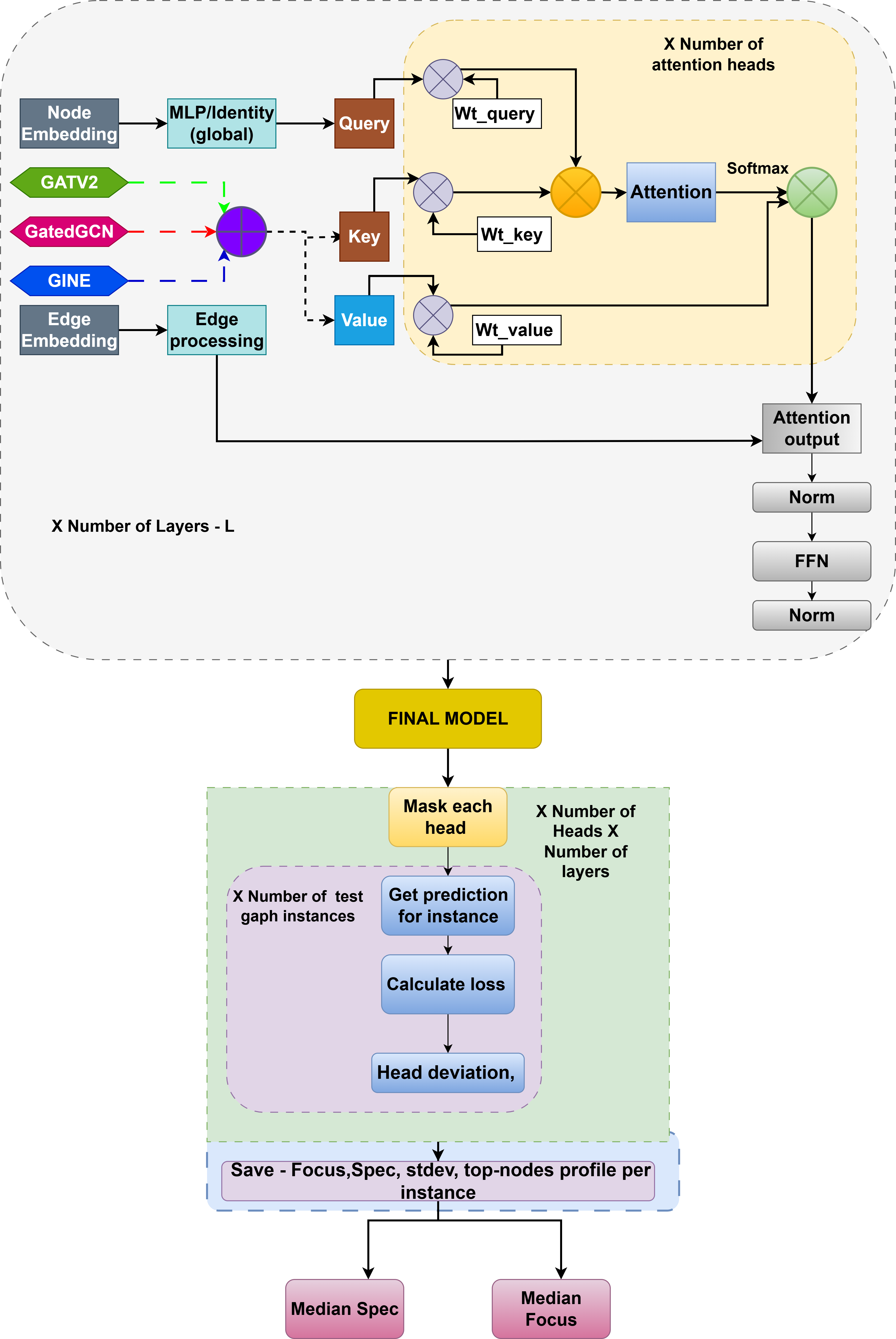}
    \caption[Asymmetric DARTS-searched GNN$\rightarrow$GT block and head-level interpretability]{
  \textbf{DARTS-GT with interpretability  framework.}
  Node features produce $Query$ via a global MLP/identity, while $Key,Value$ come from a \emph{single} GNN applied once per layer and \emph{shared across all heads}; heads differ only in their per-head linear projections, reducing cost versus per-head GNNs.
  The violet disc denotes the DARTS \emph{mixed operation} over candidate GNNs (GINE, GATv2, GatedGCN) during search; \emph{after search, exactly one GNN is fixed per layer} (e.g. final searched network in the supplementary).
  \emph{Grey discs} indicate the \emph{per-head linear transformations} (matrix multiplications with learned weights). 
  The yellow $\times$ is the score computation $QueryKey^{\top}$ (followed by Softmax), and the green $\times$ is the matrix product of the \textit{Softmax expression} with $Value$ yielding the attention output.
  An \emph{edge pathway} (edge embedding $\rightarrow$ edge processing) supplies an optional gated residual added before \textit{Norm}--\textit{FFN}--\textit{Norm}.
  \textbf{Interpretability (bottom):} on each test graph, mask one head at a time and recompute loss to obtain \emph{Head Deviation}; from these, compute per-instance \emph{Specialization} (spread across heads) and \emph{Focus} (agreement of top-$k$ heads on top-$r\%$ nodes), then report dataset medians.
  }
    \label{fig:dartsgt}
\end{figure}

\subsection{DARTS-GT}

\begin{problem}[Layer-wise Operation Selection]
For a Graph Transformer with $L$ layers and a set of candidate operations $\mathcal{O} = \{o_1, o_2, ..., o_n\}$, where $|\mathcal{O}| = n$, the goal is to find the optimal sequence of layer-wise operations from the search space $\mathcal{S}$ of $n^L$ possible configurations leading to the optimal architecture $\mathcal{A}^*$, an \textit{ordered sequence of layer-wise operations} (one per layer):
\begin{equation}
\mathcal{A}^* = \{o^{(1)*}, o^{(2)*}, ..., o^{(L)*}\}
\end{equation}
Here $o^{(\ell)*} \in \mathcal{O}$ denotes the selected operation at layer $\ell$. Formally, this architecture search is formulated as:
\begin{equation}
\label{eq:11}
\mathcal{A}^* = \arg\min_{\mathcal{A} \in \mathcal{S}} \mathcal{L}_{\text{val}}(w^*(\mathcal{A}), \mathcal{A})
\end{equation}
subject to:
\begin{equation}
\label{eq:12}
w^*(\mathcal{A}) = \arg\min_{w} \mathcal{L}_{\text{train}}(w, \mathcal{A})
\end{equation}
where $\mathcal{S}$ denotes the search space of all $n^L$ architectures, $w$ represents the network weights, and $w^*(\mathcal{A})$ are the optimal weights for architecture $\mathcal{A}$.  The following sections describe the approach we chose to solve the architecture search problem and the set of candidate operations, and hence, the search space.
\end{problem}

\subsubsection{Bilevel Optimization via DARTS}
Enumerating all possible configurations in $\mathcal{S}$ is generally infeasible, so we adopt the Differential Architecture Search (DARTS)~\cite{liu_darts_2019} framework to solve the optimization problem formulated by equations~\ref{eq:11} and~\ref{eq:12}. DARTS reformulates discrete selection as a continuous, differentiable optimization. Specifically, it introduces a softmax relaxation over candidate operations, 
casting the architecture search as the following bilevel optimisation problem~\cite{anandalingam_hierarchical_1992}:

\begin{equation}
\begin{aligned}
\min_{\alpha} \quad & \mathcal{L}_{\text{val}}(w^*(\alpha), \alpha,\mathcal{D}_{\text{DARTSval}}) \\
\text{s.t.} \quad & w^*(\alpha) = \arg\min_{w} \mathcal{L}_{\text{train}}(w, \alpha,\mathcal{D}_{\text{DARTStrain}})
\end{aligned}
\end{equation}

where $w$ represents the network weights and $\alpha = \{\alpha_o^{(\ell)} \mid o \in \mathcal{O}, \ell \in \{1, ..., L\}\}$ are continuous architecture parameters, with $\alpha_o^{(\ell)} \in \mathbb{R}$ representing the architecture weight (importance factor) for the operation $o$ at layer $\ell$.
Rather than solving the full bilevel optimization, we adopt the efficient first-order approximation of DARTS~\cite{liu_darts_2019}, which avoids Hessian–vector products. 
In practice, we first randomly shuffle the entire training dataset $\mathcal{D}_{\text{train}}$ and partition it into darts-training split, $\mathcal{D}_{\text{DARTStrain}}$ for updating weights $w$ and darts-validation split, $\mathcal{D}_{\text{DARTSval}}$ for updating architecture parameters $\alpha$, alternating the single gradient steps~\cite{liu_darts_2019}. More details about this approach can be found in~\cite{liu_darts_2019}.  Although the second-order DARTS update is more accurate, it requires expensive Hessian–vector products and offers limited practical gains in our setting.

\subsubsection{Search Space Definition}
In our work, we use the following set of candidate operations:
$\mathcal{O}$ = \{\text{GINE} \cite{wang_equivariant_2022}, \text{GATv2} \cite{brody_how_2022}, \text{GatedGCN} \cite{bresson_residual_2018}\}. Hence, $|\mathcal{O}| = n=3$ leads to a search space of $3^L$ architectures.

The following operators were selected because they represent very distinct message-passing paradigms that capture complementary aspects of graph structure.
\begin{itemize}
    \item \textbf{GINE}: Extends GIN~\cite{xu_how_2019} with edge features via permutation-invariant MLP aggregation,
    \item \textbf{GATv2}: Employs dynamic, edge-aware attention-based aggregation,
    \item \textbf{GatedGCN}: Uses edge-gated message propagation with soft edge selection. 
\end{itemize}
This diversity ensures that our search space covers complementary inductive biases while remaining computationally efficient.

\subsubsection{Asymmetric Attention Design}

Departing from the standard transformer formalism outlined in the Preliminaries section, the novelty of our approach includes an \textit{asymmetric design}: queries are derived directly from node features, while keys and values are produced by layer-specific GNN operators. This is initially motivated by the GMH design of GMT~\cite{baek_accurate_2021}, where they demonstrated that decoupling the computational sources for the attention components enhances graph-aware representations. In GMT, node features are first transformed by a GNN to obtain $H$, with queries taken directly as $Q=H$ and each head computing its own keys and values via an additional GNN applied to $H$. This asymmetry separates structural encoding from feature representation, but at the cost of computational inefficiency, since every key and value inside every head requires an independent GNN. In contrast, our design takes node features directly into $Q$, and shares the GNN operator across heads within a layer, producing an efficient formulation that still preserves the benefits of feature–structure decoupling. More broadly, rather than augmenting Graph Transformers externally with auxiliary MPNNs, a widely practiced approach \cite{rampa_recipe_nodate,song_towards_2025}, we demonstrate that graph awareness can emerge from  an architectural redesign of the attention mechanism itself.

\begin{align}
\mathbf{Q}_m^{(\ell)} &= \phi(\mathbf{Z}^{(\ell)}) \mathbf{W}_{Q,m}^{(\ell)} \\
\mathbf{K}_m^{(\ell)} &= o^{(\ell)}(\mathbf{Z}^{(\ell)}, E) \mathbf{W}_{K,m}^{(\ell)} \\
\mathbf{V}_m^{(\ell)} &= o^{(\ell)}(\mathbf{Z}^{(\ell)}, E) \mathbf{W}_{V,m}^{(\ell)}
\end{align}

where $\phi$ is a learned transformation (implemented as a two-layer MLP when additional expressiveness is beneficial, or identity otherwise) and $o^{(\ell)}$ is the operator (GNN variant) at layer $\ell$ and $o^{(\ell)}(\mathbf{Z}^{(\ell)},E)\in\mathbb{R}^{n\times d}$ ; $\mathbf{W}_{Q,m}^{(\ell)},\mathbf{W}_{K,m}^{(\ell)},\mathbf{W}_{V,m}^{(\ell)}\in\mathbb{R}^{d\times d_m}$. This design decouples the query space from the structural encoding, allowing the attention mechanism to learn optimal feature-structure alignments through the query-key interaction.

\subsubsection{Continuous relaxation (DARTS).}
During the search phase, DARTS~\cite{liu_darts_2019} relaxes the selection of discrete operators into a continuous mixture. At each layer $\ell$, instead of selecting a single operator, we compute a weighted combination:
\begin{equation}
\overline{o}^{(\ell)}(\mathbf{Z}^{(\ell)}, E) = \sum_{o \in \mathcal{O}} \frac{\exp(\alpha_o^{(\ell)})}{\sum_{o' \in \mathcal{O}} \exp(\alpha_{o'}^{(\ell)})} \cdot o(\mathbf{Z}^{(\ell)}, E)
\end{equation}

where:
\begin{itemize}
    \item $\overline{o}^{(\ell)}$ denotes the \emph{mixed} (continuous) operator at layer $\ell$, distinguished from the final discrete selection,
    \item $\alpha_o^{(\ell)} \in \mathbb{R}$ is the learnable architecture parameter for operator $o$ at layer $\ell$,
    \item $o(\mathbf{Z}^{(\ell)}, E)$ is the output of applying operator $o$ to the node features $\mathbf{Z}^{(\ell)}$ and edge set $E$.
\end{itemize}

\paragraph{Use in attention during search.}
During search, the mixed operator output $\overline{o}^{(\ell)}(\mathbf{Z}^{(\ell)},E)$ provides the structural path for the keys and values:
\begin{subequations}
\begin{align}
\mathbf{K}_{m}^{(\ell)} &= \overline{o}^{(\ell)}(\mathbf{Z}^{(\ell)},E)\,\mathbf{W}_{K,m}^{(\ell)},\\
\mathbf{V}_{m}^{(\ell)} &= \overline{o}^{(\ell)}(\mathbf{Z}^{(\ell)},E)\,\mathbf{W}_{V,m}^{(\ell)}.
\end{align}
\end{subequations}
We compute $\overline{o}^{(\ell)}(\mathbf{Z}^{(\ell)},E)$ once per layer and share it across all heads to form $\mathbf{K}_{m}^{(\ell)}$ and $\mathbf{V}_{m}^{(\ell)}$, $m \in [1,\dots, M]$.
After discretization, $\overline{o}^{(\ell)}$ is replaced by the selected operator $o^{(\ell)*}$(see $\textit{5)}$ below).
Attention may be dense or sparse, as that only changes the softmax normalization (all-pairs vs. edge-softmax) and does not alter the source of $\mathbf{K}_{m},\mathbf{V}_{m}$.

\subsubsection{Discretization and Final Architecture Training}
\label{subsection:discrete}
Upon search completion, we derive the discrete architecture by selecting the operator with maximum architecture weight at each layer:

\begin{equation}
o^{(\ell)*} = \arg\max_{o \in \mathcal{O}} \alpha_o^{(\ell)}
\end{equation}

Following the standard DARTS protocol, the final discretized architecture $\mathcal{A}^* = \{o^{(1)*}, ..., o^{(L)*}\}$ is then re-trained from scratch using the full training data. In our implementation, we instantiate a fresh model containing only the selected operations per layer. The algorithm is discussed in Algorithm~\ref{algo:1}.

\begin{algorithm}
\caption{DARTS-GT: Two-Phase Architecture Search}
\label{algo:1} 
\begin{algorithmic}[1]
\Require Graph dataset $\mathcal{D}$, candidate operations $\mathcal{O}$, $L$ layers
\Ensure Optimized architecture $\mathcal{A}^*$ with trained weights

\State \textbf{Phase 1: Architecture Search (first-order) with Mixed Operations}
\State Randomly partition $\mathcal{D}_{\text{train}}$ into $\mathcal{D}_{\text{DARTStrain}}$ (60\%) and $\mathcal{D}_{\text{DARTSval}}$ (40\%)
\State Initialize architecture parameters to zeroes, $\alpha = \{\alpha_o^{(\ell)}\}$ for all $o \in \mathcal{O}$, $\ell \in [L]$

\State Initialize network weights $w$
\For{epoch $= 1$ to $E_{\text{search}}$}
    \State Sample mini-batch from $\mathcal{D}_{\text{DARTStrain}}$ and $\mathcal{D}_{\text{DARTSval}}$
    \State Update $w$ by $\nabla_w \mathcal{L}_{\text{train}}(w, \alpha)$ \Comment{First-order step (no Hessian terms)}
\State Update $\alpha$ by $\nabla_\alpha \mathcal{L}_{\text{val}}(w, \alpha)$ \Comment{First-order step}
\Comment{Single gradient step}
\EndFor
\State \textbf{Extract optimal operations:}
\For{each layer $\ell \in [L]$}
    \State $o^{(\ell)*} \gets \arg\max_{o \in \mathcal{O}} \alpha_o^{(\ell)}$
\EndFor
\State $\mathcal{A}^* \gets \{o^{(1)*}, o^{(2)*}, ..., o^{(L)*}\}$

\State \textbf{Phase 2: Discrete Architecture Training}
\State Instantiate fresh model $\mathcal{M}_{\text{discrete}}$ with architecture $\mathcal{A}^*$
\State Train $\mathcal{M}_{\text{discrete}}$ on full $\mathcal{D}_{\text{train}}$ for $E_{\text{final}}$ epochs
\State \Return Trained model $\mathcal{M}_{\text{discrete}}$
\end{algorithmic}
\end{algorithm}

\subsubsection{Edge Residual Stream}

We inject a gated edge term into the post-attention residual (before the first normalization). For edge features $\mathbf{e}_{(i,j)}\!\in\!\mathbb{R}^{d_e}$:
\begin{align}
\tilde{\mathbf{e}}_{(i,j)} &= \mathrm{MLP}_{\text{edge}}\!\left(\mathbf{e}_{(i,j)}\right),\\
\mathbf{h}_j^{\text{edge}} &= \sum_{(i,j)\in E} \tilde{\mathbf{e}}_{(i,j)},\\
\mathbf{H}^{\text{edge}} &= \big[\,\mathbf{h}_1^{\text{edge}}\ \cdots\ \mathbf{h}_n^{\text{edge}}\,\big]^{\top}\!\in\!\mathbb{R}^{n\times d},\\
\mathbf{Y}^{(\ell)} &\gets  \mathbf{Y}^{(\ell)} + \sigma_g\!\big(\gamma^{(\ell)}\big)\,\mathbf{H}^{\text{edge}}
\end{align}

$\mathrm{MLP}_{\text{edge}}:\mathbb{R}^{d_e}\!\to\!\mathbb{R}^{d}$ is a two-layer MLP with dropout; $\sigma_g(x)$ is the sigmoid function, and $\gamma^{(\ell)}\!\in\!\mathbb{R}$ is a per-layer learnable scalar controlling the gate; $\mathbf{Y}^{(\ell)}$ is the pre-residual attention output. $\mathrm{MLP}_{\text{edge}}$ is shared across layers and $\mathbf{H}^{\text{edge}}$ is computed once per forward pass. Please see Supplementary Table I for ablation studies related to the edge residual stream.
This edge residual stream operates independently of the selected GNN operations, allowing the model to leverage edge information even when the discretized architecture uses operations with limited utilization of edge features.

\subsubsection{Computational Cost}

Our proposed asymmetric block applies a \emph{single} GNN per layer to produce the structural representations used in both $K$ and $V$. In contrast, GMT’s Graph Multi-head Attention (GMH) \cite{baek_accurate_2021} instantiates \emph{per-head} GNNs for both keys and values, i.e. $\{\mathrm{GNN}^{K}_i,\mathrm{GNN}^{V}_i\}_{i=1}^M$. Consequently, the GNN forward cost in GMH scales as $\mathcal{O}(2M \cdot \text{cost}_\text{GNN})$, while in our design it is $\mathcal{O}(1 \cdot \text{cost}_\text{GNN})$ per layer, where $\text{cost}_\text{GNN}$ denotes the computational cost of a single GNN forward pass per layer. Thus, we preserve the feature–structure interaction ($Q$ from features; $K,V$ from GNN) while reducing the structural computation across heads.
From the perspective of architecture search overhead, DARTS~\cite{liu_darts_2019} uses a first-order approximation that avoids Hessian–vector product overhead.

\subsection{Interpretability Framework}
\label{method:interpret}

\subsubsection{Framework Overview}
We propose a systematic framework to quantify Graph Transformer interpretability through head-level and node-level analysis. Our approach goes beyond visualization and yields \emph{quantitative}, \emph{instance-specific} evidence of which components \emph{causally} drive the predictions. We introduce \textit{Head-Deviation}, \textit{Specialization}, and \textit{Focus}, aligned with established interpretability principles where \emph{fidelity}, \emph{sparsity}, and \emph{consensus} are standard evaluation criteria~\cite{amara_graphframex_2024,lukyanov_robustness_2025}. Because these metrics are ablation-based, they measure causal contribution rather than correlation, allowing robust, fair comparisons of interpretability across architectures.

\paragraph{Masking protocol (causal ablation).}
For head $m$ in layer $\ell$, we implement a weight-level ablation that removes its contribution at inference. Concretely, we \emph{silence head $m$ in layer $\ell$, by zeroing its pre-concatenation output} $Y_m^{(\ell)}$ (equivalently, zeroing the corresponding slice of $\mathbf{W}_{\text{out}}$ in Eq.~\ref{eq:mh-concat}). All other parameters, including the outputs of other heads, layer norms, and feed-forward sublayers, are left unchanged. The models are evaluated in \texttt{eval} mode (no dropout), with a fixed seed and without any retraining or adaptation.

\subsubsection{Head Deviation}
To quantify the importance of individual attention heads, we measure how their removal affects the task loss. For a graph $G_i$ with the true label $y_i$, let $f(G_i)$ denote the prediction of the model and $f^{-m,\ell}(G_i)$ denote the prediction with head $m$ in layer $\ell$ masked.

\begin{definition}[Head Deviation]
The head deviation measures the change in task loss when head $m$ in layer $\ell$ is masked:
\begin{equation}
\delta_{m,\ell,i} \;=\; \mathrm{loss}\!\left(f^{-m,\ell}(G_i),\, y_i\right) \;-\; \mathrm{loss}\!\left(f(G_i),\, y_i\right),
\end{equation}
\end{definition}

Here, $\mathrm{loss}(\cdot,\cdot)$ denotes the loss appropriate to the machine learning task (e.g., MAE for regression, binary cross-entropy for binary classification, cross-entropy for multi-class classification). We quantify head importance through \textit{causal intervention}: the deviation $\delta_{m,\ell,i}$ measures how the prediction changes when head $m$ in layer $\ell$ is ablated, providing direct evidence of its functional necessity rather than mere correlational patterns. This approach asks "\textit{what happens without this head?}" , distinguishes truly important components from those that merely appear salient in visualizations.

\paragraph{Head ranking}
For each instance, we rank the heads by their deviation $\delta_{m,\ell,i}$ in descending order. 
\emph{Beneficial heads} ($\delta_{m,\ell,i}>0$, masking increases loss) thus appear first, while \emph{harmful heads} ($\delta_{m,\ell,i}<0$) appear last. 
When we need a sign-agnostic view, we report the magnitude $|\delta_{m,\ell,i}|$ as a measure of head impact.

\subsubsection{Global Interpretability Metrics}
Beyond individual head importance, we summarize \emph{how interpretable} the model is for a given instance across all heads. We define two complementary metrics: \textit{specialization} (importance concentration or head-distinguishability) and \textit{focus} (consensus of important heads on nodes).

\begin{definition}[Specialization]
For graph $G_i$,
\begin{equation}
\mathrm{Spec}(G_i) \;=\; \mathrm{std}\Big(\{\delta_{m,\ell,i}\}_{m \in \mathcal{M},\, \ell \in \mathcal{L}}\Big),
\end{equation}
where $\mathrm{std}(.)$ denotes the standard deviation across all heads' deviations with $\mathcal{M} = \{1, . . ., M\}$, $\mathcal{L} = \{1, . . .,L\}$ where $M$ is the number of attention-heads and $L$ total number of layers, defined earlier.

\end{definition}

High specialization indicates that heads are easily distinguishable by their causal impact, whereas low specialization suggests similar head impacts, making it harder to isolate drivers unless complemented by the focus metric.

\begin{definition}[Focus] 

Let $\mathcal{P}_i^* = \{p_1,\dots,p_k\}$ be the top-$k$ head-layer pairs for instance $G_i$, ranked according to the head-ranking rule defined above, where $p_j=(i'_j,i''_j)$, $j=1,...,k$, $i'_j \in \mathcal{M}$, $i''_j \in \mathcal{L}$ and $i'_j$, $i''_j$ are index variables.  For a pair $p$, we define $N_{p}$ as the set of \emph{top-$r$\% nodes by incoming attention mass} for that head on $G_i$. $N_{p}$ contains the {top-$r$}\%  nodes by the following values:
\begin{itemize}
    \item \textbf{Dense attention:} with head softmax matrix $\mathbf{S}^{(\ell)}_{m}$, incoming mass for node $j$ is $\sum_{i} [\mathbf{S}^{(\ell)}_{m}]_{ij}$;
    \item \textbf{Sparse attention:} with edge-softmax $s_{m,ij}$ on $(i,j)\in E$, incoming mass for node $j$ is $\sum_{i:(i,j)\in E} s_{m,ij}$.  
\end{itemize}
The Focus metric computes the average Jaccard similarity between all unique pairs:
\begin{equation}
\text{Focus}(G_i) \;=\; \frac{1}{\binom{k}{2}} \sum_{j=1}^{k-1} \sum_{m=j+1}^{k} 
\frac{|N_{p_j} \cap N_{p_m}|}{|N_{p_j} \cup N_{p_m}|},
\end{equation}
where $\binom{k}{2} = \tfrac{k(k-1)}{2}$ is the number of unique head pairs.
\end{definition}

Unless stated otherwise, in our implementation, we use $k{=}5$ and a $r=10\%$ node fraction. This metric quantifies the degree of consensus among the most causally important heads, i.e. measure top-k-head agreement on top-$r\%$
nodes. A high Focus score indicates that the model's key components agree on the same salient graph substructures, simplifying interpretation.

\subsubsection{Dataset-Level Aggregation}
For dataset $\mathcal{D}$ with $N$ graphs, we use the median for robustness to instance variability and outliers:
\begin{align}
\overline{\mathrm{Spec}} &= \mathrm{median}_{i=1}^{N} \{\mathrm{Spec}(G_i)\}, \\
\overline{\mathrm{Focus}} &= \mathrm{median}_{i=1}^{N} \{\mathrm{Focus}(G_i)\}.
\end{align}

\subsubsection{Interpretation Guidelines}
The combination of specialization and focus determines how interpretable an instance-level prediction is.
\begin{itemize}
    \item \textbf{High Spec + High Focus}: Most interpretable. Few heads dominate and agree on important nodes.
    \item \textbf{High Spec + Low Focus}: Important heads are identifiable but attend to different nodes, suggesting complementary strategies.
    \item \textbf{Low Spec + High Focus}: Uniform head importance but consensus on nodes. Clear node-level attribution despite unclear head roles.
    \item \textbf{Low Spec + Low Focus}: Least interpretable. Uniform importance with disparate attention patterns.
\end{itemize}
$\mathrm{Focus}\in[0,1]$ measures consensus on node attribution. $\mathrm{Spec}\in[0,\infty)$ measures concentration of head importance or head-distinguishability. Together, they reveal which heads matter (specialization) and what they attend to (focus), providing faithful, comparable, instance-level interpretability and a principled route to model-level comparison. The complete process is depicted in Algorithm~\ref{alg:interpretability}.

\begin{algorithm}[t]
\caption{Graph Transformer Interpretability Analysis}
\label{alg:interpretability}
\begin{algorithmic}[1]
\Require Model $f$, Test set $\mathcal{D}_{\text{test}}$, Top heads $k$
\Ensure $\{\mathrm{Spec}(G_i), \mathrm{Focus}(G_i)\}$ for each $G_i$
\State \textbf{Phase 1: Compute Head Deviations}
\For{$\ell = 1$ to $L$}
    \For{$m = 1$ to $M$}
        \State $f^{-m,\ell} \gets$ model with head $m$ in layer $\ell$ masked
        \For{$G_i \in \mathcal{D}_{\text{test}}$}
            \State $e_{\text{base}} \gets \mathrm{loss}\!\left(f(G_i), y_i\right)$
            \State $e_{\text{masked}} \gets \mathrm{loss}\!\left(f^{-m,\ell}(G_i), y_i\right)$
            \State $\delta_{m,\ell,i} \gets e_{\text{masked}} - e_{\text{base}}$
        \EndFor
    \EndFor
\EndFor
\State \textbf{Phase 2: Compute Metrics}
\For{$G_i \in \mathcal{D}_{\text{test}}$}
    \State $\Delta_i \gets \{\delta_{m,\ell,i} : m=1..M,\; \ell=1..L\}$
    \State $\mathrm{Spec}(G_i) \gets \mathrm{std}(\Delta_i)$
    \State $\mathcal{P}^*_i \gets$ top-$k$ pairs of head $m$ in layer $l$: $(m,\ell)$ ranked by $\delta_{m,\ell,i}$ (descending)
    \For{$p \in \mathcal{P}^*_i$}
        \State $N_{p_i} \gets$ top-$r\%$ nodes by incoming attention mass for $p$ on $G_i$ 
    \EndFor
    \State $\mathrm{Focus}(G_i) \gets \frac{1}{\binom{k}{2}}
        \sum_{j=1}^{k-1}\sum_{m=j+1}^{k}
        \frac{|N_{p_j} \cap N_{p_m}|}{|N_{p_j} \cup N_{p_m}|}$
\EndFor
\State \Return $\{(\mathrm{Spec}(G_i), \mathrm{Focus}(G_i)) : G_i \in \mathcal{D}_{\text{test}}\}$
\end{algorithmic}
\end{algorithm}

\section{Experimental Results}

In this section, we comprehensively evaluate DARTS-GT in multiple aspects to establish both its performance advantages and its interpretability characteristics. Our experimental analysis comprises five key components: 
\begin{enumerate}
\item Direct comparison with standard (vanilla) Graph Transformer~\cite{dwivedi_generalization_2021} to isolate the impact of our architectural innovations; 
\item Comparison against state-of-the-art baselines including hand-designed GNNs, recent Graph Transformers, and neural architecture search methods;   
\item Analysis of heterogeneous versus uniform architectures and random-search obtained architectures, to validate the benefits of layer-wise GNN diversity through DARTS search; 
\item Ablation studies examining the role of asymmetric design while understanding edge features role is given in the supplementary; 
\item Quantitative interpretability analysis using our novel framework to assess model transparency. Our evaluation of the interpretability framework operates on two levels.
    \begin{enumerate}
        \item  At the population level, we quantify overall model interpretability through metrics (Specialization, Focus) that characterize potentially how easily practitioners can identify important heads and nodes driving predictions. 
        \item  At the instance level, we demonstrate how our framework reveals specific graph components critical for individual predictions.
    \end{enumerate}
\end{enumerate}

To ensure a robust evaluation, we perform experiments with 4 random seeds for SOTA baseline comparisons (Tables III-IV) and vanilla-GT versus DARTS-GT comparisons (Table II) following the benchmarking protocol introduced in \cite{dwivedi_benchmarking_2022}, while using 3 seeds for architectural analysis experiments (Tables V-VII,VIII) due to computational constraints, which is also widely followed protocol \cite{rampa_recipe_nodate}.

\subsection{Dataset and Task Selection}

We evaluated DARTS-GT against three categories of baselines: SOTA GNN architectures (e.g. GCN~\cite{kipf_semi-supervised_2017}, GIN~\cite{xu_how_2019}, GatedGCN~\cite{bresson_residual_2018}, PNA~\cite{corso_principal_2020}), recent Graph Transformer models (e.g. SAN~\cite{kreuzer_rethinking_2021}, Graphormer~\cite{ying_transformers_2021}, GPS~\cite{rampa_recipe_nodate}), and neural architecture search methods for Graph Transformers (UGAS~\cite{song_towards_2025}, EGTAS~\cite{wang_automatic_2024}, AutoGT~\cite{zhang_autogt_2023}). Our evaluation spans eight diverse datasets that encompasses both graph-level and node-level prediction tasks, as detailed in Table~\ref{tab:dataset_stats}.

\begin{table}[!h]
\centering
\caption{Dataset Statistics and Characteristics}
\label{tab:dataset_stats}
\resizebox{\columnwidth}{!}{%
\begin{tabular}{lrrrlc}
\toprule
Dataset & \#Graphs & Avg\# & Avg\# & Prediction & Task \\
 & & Nodes & Edges & Level &  \\
\midrule
ZINC & 12,000 & 23.2 & 24.9 & Graph & Regression \\
OGBG-MolHIV & 41,127 & 25.5 & 27.5 & Graph & Binary Cls. \\
OGBG-MolPCBA & 437,929 & 26.0 & 28.1 & Graph & Multi-task Cls. \\
MalNet-Tiny & 5,000 & 1,410.3 & 2,859.9 & Graph & Multi-class Cls. \\
PATTERN & 14,000 & 118.9 & 3,039.3 & Node & Binary Cls. \\
CLUSTER & 12,000 & 117.2 & 2,150.9 & Node & Multi-class Cls. \\
Peptides-Struct
 & 15,535 & 150.9 & 307.3 & Graph & Multi-task Reg. \\
Peptides-Func
 & 15,535 & 150.9 & 307.3 & Graph & Multi-task Cls. \\
\bottomrule
\end{tabular}%
}
\vspace{-3mm}
\end{table}

The selected benchmarks~\cite{rampa_recipe_nodate} provide comprehensive coverage in multiple dimensions. This range accommodates diverse tasks such as graph regression (single and multi-task), graph classification (binary, multi-class, and multi-task), and node-level predictions (binary and multi-class classifications). The datasets exhibit substantial variation in scale, from small molecular graphs in ZINC (averaging 23 nodes) to large-scale graphs in MalNet-Tiny (averaging 1,410 nodes), testing the ability of our architecture to handle varying structural complexities. Dataset sizes range from 5,000 to more than 437,000 graphs, ensuring robust evaluation across different dataset sizes. This diversity, spanning molecular property prediction (OGBG), biological sequences (Peptides), and synthetic benchmarks (PATTERN, CLUSTER), demonstrates that our architectural choices generalize beyond domain-specific patterns. Notably, the inclusion of two Long-Range Graph Benchmark (LRGB) datasets, Peptides-Struct and Peptides-Func, facilitated evaluation of the capacity of our model to capture long-range dependencies, a critical challenge for Graph Transformers.

\subsection{Comparison Against Standard Graph Transformers}

Table~\ref{tab:attention_comparison} presents the performance comparison of DARTS-GT with the standard Graph Transformer (Vanilla-GT) introduced by Dwivedi et al. \cite{dwivedi_generalization_2021}, which forms the foundation of many subsequent works \cite{song_towards_2025}. Unlike recent approaches that enhance graph transformers through parallel MPNN paths \cite{rampa_recipe_nodate}, novel positional encoding \cite{kreuzer_rethinking_2021}, or structural encoding-augmented attention \cite{ying_transformers_2021}, DARTS-GT alters the core attention mechanism through asymmetric query-key value computation and heterogeneous per-layer GNN selection. This is a significant departure from standard GT where Q, K, V share the same computational source without GNNs. This experiment quantifies the improvement of DARTS-GT  over the baseline it extends.
For rigorous comparison, we standardized most experimental conditions using the GPS configuration \cite{rampa_recipe_nodate}. We extend our implementation from their opensource repository, adopting their fixed positional encoding, hidden dimensions, attention heads, dropout rates, and pre/post-processing components for both models. Only the layer count was adjusted to maintain a comparable parameter budget range. This controlled setup (Supplementary-Tables II,III,IV), prioritizing methodological clarity over peak performance, ensures that observed improvements stem from design changes rather than hyperparameter tuning or auxiliary components.

\renewcommand{\arraystretch}{1.5}  
\begin{table}[!h]
\centering
\Huge        
\caption{Performance Comparison: Vanilla-GT vs DARTS-GT across diverse datasets with Dense and Sparse Attention Mechanisms. $\Delta$ represents the relative improvement of DARTS-GT over Vanilla-GT, calculated as $\frac{|\text{DARTS-GT} - \text{Vanilla-GT}|}{|\text{Vanilla-GT}|} \times 100\%$, where improvement direction accounts for metric type (decrease for MAE, increase for others). Shown is the mean ± std across 4 seeds. Best performances are highlighted in \textbf{bold}.}
\label{tab:attention_comparison}
\resizebox{\columnwidth}{!}{%
\begin{tabular}{llccccc} 
\toprule
\textbf{Attn.} & \textbf{Model} & \textbf{Pep-Struct} & \textbf{Pep-Func} & \textbf{PATTERN} & \textbf{MolHIV} & \textbf{MolPCBA} \\
& & MAE$\downarrow$ & AP$\uparrow$ & Acc$\uparrow$ & AUROC$\uparrow$ & AP$\uparrow$ \\
\midrule
\multirow{3}{*}{Dense} 
& Vanilla-GT & 0.254±.002 & 0.623±.009 & 0.678±.164 & 0.761±.009 & 0.048±.007 \\
& \textbf{DARTS-GT } & \textbf{0.247±.001} & \textbf{0.669±.004} & \textbf{0.852±.001} & \textbf{0.766±.016} & \textbf{0.201±.008} \\
& $\Delta$(\%) & 2.8\% & 7.4\% & 25.7\% & 0.7\% & 318\% \\
\midrule
\multirow{3}{*}{Sparse} 
& Vanilla-GT & 0.268±.001 & 0.557±.005 & 0.681±.199 & 0.758±.009 & 0.241±.032 \\
& \textbf{DARTS-GT } & \textbf{0.263±.001} & \textbf{0.609±.014} & \textbf{0.867±.000} & \textbf{0.776±.006} & \textbf{0.258±.004} \\
& $\Delta$(\%) & 1.9\% & 9.3\% & 27.3\% & 2.4\% & 7.1\% \\
\bottomrule
\end{tabular}%
}
\vspace{-2mm}
\end{table}

We tested both dense (full attention) and sparse attention mechanisms to understand how architectural improvements translate across different computational settings. The results reveal consistent gains across all benchmarks, with particularly striking improvements on PATTERN (25-27\% relative accuracy increase) and OGBG-MolPCBA (318\% relative improvement in average precision for dense attention). Even on long-range benchmarks (Peptides), where standard transformers already perform reasonably, DARTS-GT achieves meaningful non-trivial improvements over the baseline. Our comparison reveals that standard Graph Transformers, when isolated from auxiliary components, exhibit significant limitations, particularly evident in MolPCBA dense attention. This raises a critical question about hybrid architectures like GPS (which uses dense attention with parallel MPNN): do performance gains stem primarily from the MPNN path itself, or from the synergistic interaction where MPNN features and transformer attention are jointly fed into subsequent layers? The DARTS-GT design removes this ambiguity by demonstrating competitive performance using only enhanced attention mechanisms without any parallel MPNN component. This isolation proves that the Graph Transformer itself can be made structurally aware through architectural innovation in the attention mechanism, rather than relying on auxiliary components or cross-pathway synergies.

\subsection{Comparison Against SOTA leaderboard}

Table~\ref{tab:SOTA1} and Table~\ref{tab:SOTA2} benchmark DARTS-GT against three categories of methods: hand-designed GNNs (e.g., GCN, GIN, GatedGCN, PNA), Graph Transformers (e.g., SAN, Graphormer, GPS), and NAS-based GT approaches (EGTAS, AutoGT, UGAS). Each method is reported under its validation-selected configuration (dense or sparse). Full settings are available in supplementary. DARTS-GT achieves state-of-the-art results on four datasets: ZINC (0.066), MalNet-Tiny (93.25\%), and both LRGB benchmarks: Peptides-Func (0.669) and Peptides-Struc (0.246). Most notably, for the other four datasets, PATTERN, CLUSTER, MolHIV, MolPCBA, there are \textit{distinct} best performing methods, namely EGTAS (for CLUSTER), UGAS (for PATTERN), CIN (for MolHIV) and CRAW1 (for MolPCBA).  \textit{Therefore, DARTS-GT substantially stands out as outperforming all methods in four datasets, while none of the other four highlighted methods or any other methods outperformed all methods for more than one dataset.}

\begin{table}[!t]
\centering
\caption{Test Performance on Standard Benchmarks. Shown is the mean ± std across 4 seeds. Best performance is \textbf{bolded} while second best is \underline{underlined}. Datasets where we have outperformed GPS (but not best) is marked with asterisks}
\label{tab:SOTA1}
\resizebox{\columnwidth}{!}{%
\begin{tabular}{l|cccc}
\textbf{Model} & \textbf{ZINC}  & \textbf{MALNET} & \textbf{PATTERN} & \textbf{CLUSTER}  \\
 & MAE $\downarrow$  & Acc $\uparrow$ & Acc $\uparrow$ & Acc $\uparrow$ \\
\midrule
GCN~\cite{kipf_semi-supervised_2017} & 0.367±0.011  & NA & 71.892±0.334 & 68.498±0.976 \\
GIN~\cite{xu_how_2019} & 0.526±0.051  & NA & 85.387±0.136 & 64.716±1.553 \\
GAT~\cite{velickovic_graph_2018} & 0.384±0.007 & NA & 78.271±0.186 & 70.587±0.447 \\
GatedGCN~\cite{wu_representing_2022} & 0.282±0.015  &  0.9223±0.65 & 85.568±0.088 & 73.840±0.326 \\
GatedGCN-LSPE~\cite{dwivedi_graph_2022} & 0.090±0.001  &NA & NA & NA \\
PNA~\cite{corso_principal_2020} & 0.188±0.004 & NA & NA & NA \\
DGN~\cite{beaini_directional_2021} & 0.168±0.003  & NA & 86.680±0.034 & NA \\
GSN~\cite{bouritsas_improving_2023} & 0.101±0.010  & NA & NA & NA \\
CIN~\cite{bodnar_weisfeiler_2022} & 0.079±0.006  & NA & NA & NA \\
CRaWl~\cite{tonshoff_walking_2023} & 0.085±0.004  & NA & NA & NA \\
GIN-AK+~\cite{zhao_stars_2022} & 0.080±0.001  & NA & \underline{86.850±0.057} & NA \\
\midrule
EGTAS~\cite{wang_automatic_2024} & 0.075±0.073  & NA & 86.742±0.053 & \textbf{79.236±0.215}\\
UGAS~\cite{song_towards_2025} & NA  & NA &  \textbf{86.89±0.02} & 78.14±0.21\\
\midrule
SAN~\cite{kreuzer_rethinking_2021} & 0.139±0.006  & NA & 86.581±0.037 & 76.691±0.65\\
Graphormer~\cite{ying_transformers_2021} & 0.122±0.006  & NA & NA & NA \\
K-Subgraph SAT~\cite{chen_structure-aware_2022} & 0.094±0.008  & NA & {86.848±0.037} & 77.856±0.104\\
EGT~\cite{hussain_global_2022} & 0.108±0.009  & NA & 86.821±0.020 & \underline{79.232±0.348}\\
GPS~\cite{rampa_recipe_nodate} & \underline{0.070±0.004}  & \underline{0.9264±0.78} & 86.685±0.059 & 78.016±0.180\\
\midrule
\textbf{DARTS-GT} & \textbf{0.066±0.003}  & \textbf{0.9325±0.006} & 86.68±0.036 & 78.299±0.07*\\
\bottomrule
\end{tabular}%
}
\end{table}

For LRGB~\cite{dwivedi_long_2023} benchmarks, DARTS-GT advances beyond both GPS and the recent UGAS method, with considerably lower variance on Peptides-Struc than the latter (±0.0006 vs UGAS's ±0.002), indicating more stable solutions. The consistent gains across diverse tasks validate that differentiable architecture search discovers effective  designs within the transformer framework. In addition to the SOTA performance achieved by the DARTS-GT discovered architectures, they exhibit enhanced interpretability characteristics, as demonstrated in sections~\ref{subsection:Interpretability1} and~\ref{subsection:Interpretability2}.

\renewcommand{\arraystretch}{1.5}
\begin{table}[!t]
\centering
\Huge
\caption{Test Performance on OGB and LRGB Benchmarks. Shown is the mean ± std across 4 seeds. Best performance is \textbf{bolded} while second best is \underline{underlined}.}
\label{tab:SOTA2}
\resizebox{\columnwidth}{!}{%
\begin{tabular}{l|cccc}
\textbf{Model} & \textbf{MolHIV} & \textbf{MOLPCBA} & \textbf{PEP-FUNC} & \textbf{PEP-STRUC} \\
& AUROC $\uparrow$ & AP $\uparrow$ & AP $\uparrow$ & MAE $\downarrow$ \\
\midrule
GCN & NA & NA & 0.5930 ± 0.002 &  0.3496 ± 0.0013 \\
GINE~\cite{wang_equivariant_2022} & NA & NA & 0.5498 ± 0.0079 &  0.3547 ± 0.0045 \\
GCN+virtualnode~\cite{rampa_recipe_nodate} & 0.7599±0.0119 & 0.2424±0.0034 & NA & NA \\
GIN+virtualnode~\cite{rampa_recipe_nodate} & 0.7707±0.0149 & 0.2703±0.0023 & NA & NA \\
GatedGCN & NA & NA &  0.5864 ± 0.0077 & 0.3420 ± 0.0013\\
GatedGCN-LSPE & NA & 0.267±0.002 & NA & NA \\
PNA & 0.7905±0.0132 & 0.2838±0.0035 & NA & NA \\
DeeperGCN~\cite{rampa_recipe_nodate} & 0.7858±0.0117 & 0.2781±0.0038 & NA & NA \\
DGN & 0.7970±0.0097 & 0.2885±0.0030 & NA & NA \\
GSN(directional) & 0.8039±0.0090 & NA & NA & NA \\
CIN & \textbf{0.8094±0.0057} & NA & NA & NA \\
CRaWl & NA & \textbf{0.2986±0.0025} & NA & NA\\
GIN-AK+ & 0.7961±0.0119 & \underline{0.2930±0.0044} & NA & NA \\
ExpC~\cite{yang_breaking_2023} & 0.7799±0.0082 & 0.2342±0.0029 & NA & NA \\
\midrule
AUTOGT & 0.7495±0.0102 & NA & NA & NA \\
EGTAS & 0.7981±0.0117 & NA & NA & NA \\
UGAS & \underline{0.7997±0.51} & NA & \underline{0.667±0.007} & \underline{0.247±0.002} \\
\midrule
SAN & 0.7785±0.2470 & 0.2765±0.0042 & NA & NA\\
SAN+LAPPE~\cite{rampa_recipe_nodate} & NA & NA & 0.6384±0.0121 & 0.2683±0.0043\\
GraphTrans~\cite{wu_representing_2022} & NA & 0.2761±0.0029 & NA & NA\\
GPS~\cite{rampa_recipe_nodate} & 0.7880±0.0101 & 0.2907±0.0028 & 0.6535±0.0041 & 0.2500±0.0005\\
\midrule
\textbf{DARTS-GT} & 0.7766±0.006 & 0.2584±0.004 & \textbf{0.669±0.004} & \textbf{0.246±0.0006}\\
\bottomrule
\end{tabular}%
}
\end{table}

\subsection{Dissecting DARTS-GT's Architecture Discovery Patterns}

\begin{figure}
    \centering
    \includegraphics[width=1\linewidth]{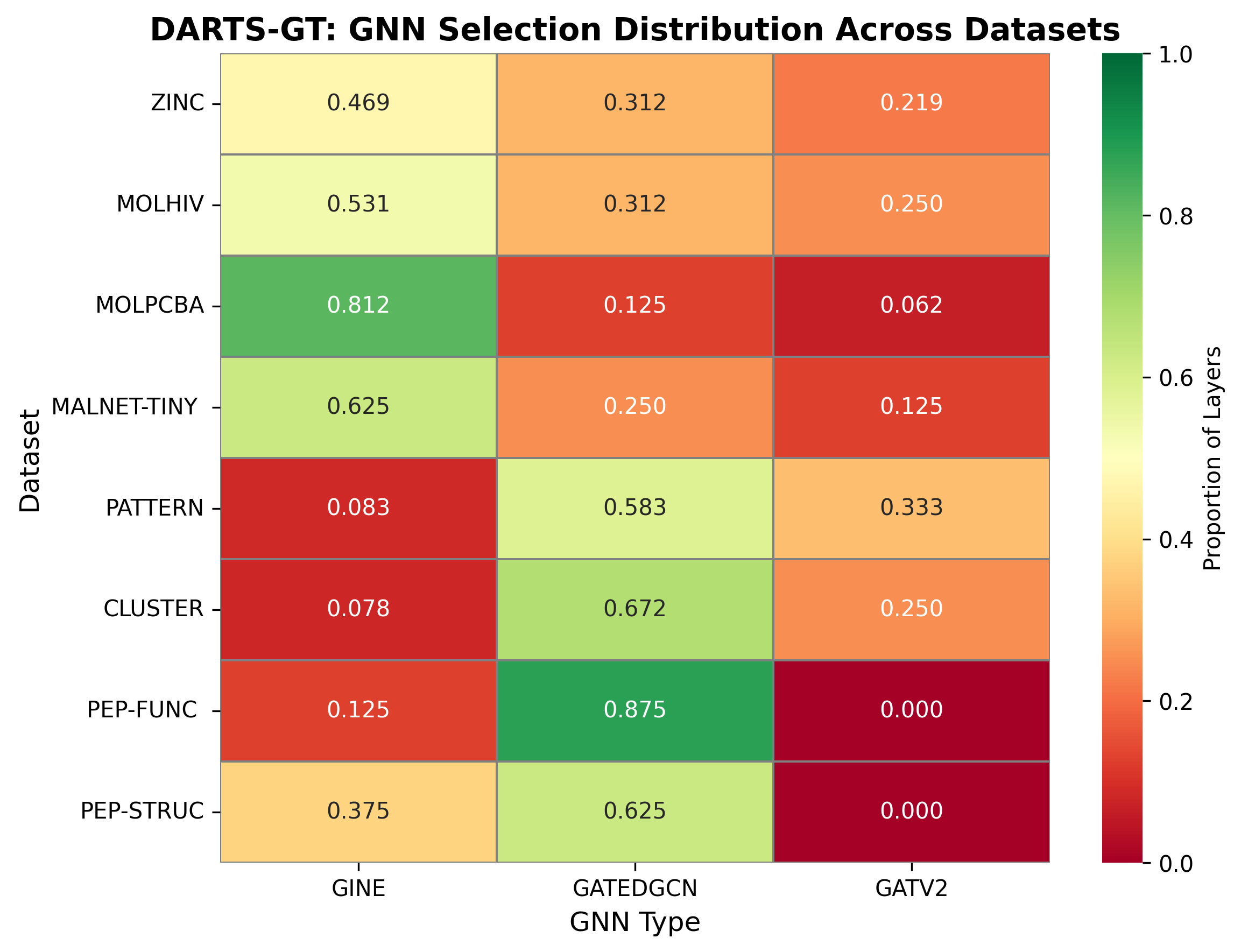}
    \caption{ Heatmap showing the proportion of each GNN type selected by DARTS-GT across different datasets. The values indicate  fraction of total layers using each GNN operation, that is averaged over 4 seeds. Color intensity reflects the usage frequency as indicated}
    \label{fig:layer_distribution}
\end{figure}

Figure~\ref{fig:layer_distribution} visualizes the GNN distributions discovered by DARTS across eight datasets, revealing prominent diversity in architectural choices. The selection patterns range from highly specialized (MolPCBA with 81\% GINE dominance) to nearly uniform distributions (ZINC, MolHIV with balanced GNN usage). This demonstrates that DARTS adapts to dataset-specific requirements rather than converging to a universal solution, as expected from a neural architecture search (NAS) algorithm. Interestingly, node-level tasks (PATTERN, CLUSTER) favored GatedGCN. We could also observe the complete absence of GATV2 in Peptides datasets, coupled with GatedGCN's dominance (87.5\% in Peptides-Func). However, these proportions only show that the GNN selection is utilised, but the layer-wise positioning of each GNN, which DARTS optimizes through gradient-based search, is equally critical. A dataset might use 50\% GINE, but whether these appear in early or late layers inherently impacts feature extraction and performance (examples of a few searched architectures are given in Supplementary Figure 2).


Although Figure~\ref{fig:layer_distribution} reveals that GatedGCN dominates in CLUSTER (67.2\%), this prevalence does not indicate superiority as a standalone operation. Table~\ref{tab:darts_vs_uniform} exposes a counterintuitive phenomenon: \textit{The most-used GNN in a heterogeneous architecture may not be the best performing uniform architecture.}

\renewcommand{\arraystretch}{1.5}
\begin{table}[!t]
\centering
\caption{Comparison of DARTS-GT against uniform (homogeneous) and random-searched architectures across different attention types and dataset variety. Shown is the mean ± std across 3 seeds. Best results are in \textbf{bold}.}
\label{tab:darts_vs_uniform}
\setlength{\tabcolsep}{4pt} 
\begin{tabular}{l|c|l|c}
\toprule
\textbf{Dataset} & \textbf{Attn-Type} 
& \textbf{Model Type} & \textbf{Perf} \\
\midrule
\multirow{5}{*}{CLUSTER} & \multirow{5}{*}{Sparse} 
    & Random & 0.7388±0.0741 \\
    & & Uniform-GINE & 0.7688±0.0018 \\
    & & Uniform-GATEDGCN & 0.7047±0.069 \\
    & & Uniform-GATV2 & 0.7805±0.00213 \\
    & & \textbf{DARTS-GT (ours)} & \textbf{0.7835±0.0004} \\
\midrule
\multirow{5}{*}{Peptides-Func} & \multirow{5}{*}{Dense}
    & Random & 0.646±0.0048 \\
    & & Uniform-GINE & 0.6555±0.0093 \\
    & & Uniform-GATEDGCN & 0.6718±0.0035 \\
    & & Uniform-GATV2 & 0.6530±0.0081 \\
    & & \textbf{DARTS-GT (ours)} & \textbf{0.6730±0.0056} \\
\midrule
\multirow{5}{*}{Malnet-Tiny} & \multirow{5}{*}{Sparse}
    & Random & 0.919±0.0081 \\
    & & Uniform-GINE & 0.919±0.0081 \\
    & & Uniform-GATEDGCN & 0.929±0.0060 \\
    & & Uniform-GATV2 & 0.922±0.0098 \\
    & & \textbf{DARTS-GT (ours)} & \textbf{0.934±0.0057} \\
\bottomrule
\end{tabular}
\end{table}

Table~\ref{tab:darts_vs_uniform} compares DARTS-GT against uniform architectures and random heterogeneous selection (mean of 3 different random configurations). Despite dominant operations in DARTS-GT architectures (e.g., GatedGCN at 67.2\% in CLUSTER, 87.5\% in Peptides-Func), uniform deployment of these same operations underperforms: uniform GatedGCN achieves 0.7047 vs DARTS-GT's 0.7835 on CLUSTER and 0.6718 vs 0.6730 on Peptides-Func. On MalNet-Tiny, uniform GatedGCN outperforms uniform GINE despite GINE dominating (62.5\%) the DARTS architecture. This demonstrates that no single operation is universally optimal, and "popular" operations in DARTS-GT architecture can underperform when deployed homogeneously.
Although DARTS-GT's gains appear modest, they align with typical graph benchmark improvements where SOTA advances are often under 1\% (as seen in SAN, GPS, UGAS and our Tables~\ref{tab:SOTA1}-\ref{tab:SOTA2}). These gains are consistent across datasets/seeds, not isolated cases. Moreover, finding even the best uniform architecture requires testing $n$ separate models, while the heterogeneous space grows exponentially ($3^4=81$
 for 4 layers; $3^{10}=59,049$  for 10 layers). DARTS-GT efficiently navigates this space without exhaustive enumeration.

Three key insights emerge: (1) DARTS discovers synergistic layer-wise combinations rather than repeating the strongest operation; (2) DARTS-GT consistently outperforms random heterogeneous architectures, confirming principled optimization beats chance; (3) DARTS-GT's value lies in removing the rigid commitment to uniformity or heterogeneity in advance. It adaptively recovers strong uniform solutions when sufficient and heterogeneous mixtures when advantageous.
Our contribution extends beyond numerical improvements. While prior GTs relied on auxiliary GNN paths or hand-designed uniform backbones, we show that the core transformer architecture itself can be systematically optimized. This reframes GT design as an open direction for principled exploration, where the demonstrated consistent improvements evidence yet unexplored directions in Graph Transformers.
\subsection{Ablation Studies: Assymetric vs Symmetric attention}

Although Table~\ref{tab:attention_comparison} demonstrated DARTS-GT's superiority over Vanilla-GT (standard GT), a critical question remains: Is asymmetric attention design essential, or would a symmetric variant where $Q_{m}$ ,$K_{m}$ and $V_{m}$ all derive from GNN outputs perform better?

To address this, we conducted controlled experiments across 3 seeds comparing our asymmetric design against its symmetric counterpart.  All other experimental conditions remained identical between models.

The results obtained consistently favored the DARTS-GT asymmetric design over the symmetric variant in all configurations tested. This validates our architectural choice: decoupling queries from structural encoding ($Q_{m}$ from features, $K_{m}$/$V_{m}$ from GNNs) provides superior performance compared to the seemingly natural approach of deriving all attention components from the same GNN source. 

\begin{table}[!h]
\centering
\caption{Ablation Study: Symmetric ($Q_{m}$ ,$K_{m}$ and $V_{m}$)  vs Asymmetric Attention Design (DARTS-GT design ).Shown is the mean ± std across 3 seeds. Best performance is \textbf{bolded}. }
\label{tab:symmetric_vs_asymmetric}
\resizebox{\columnwidth}{!}{%
\begin{tabular}{llcc}
\toprule
\textbf{Dataset} & \textbf{Attn-Type} & \textbf{Model Type} & \textbf{Performance} \\
\midrule
\multirow{2}{*}{CLUSTER} & \multirow{2}{*}{Sparse} 
& Symmetric  & 78.2±0.0010 \\
& & \textbf{DARTS-GT (Asymmetric)} & \textbf{78.35±0.0004} \\
\midrule
\multirow{2}{*}{Peptides-Func} & \multirow{2}{*}{Dense} 
& Symmetric  & 0.667±0.009 \\
& & \textbf{DARTS-GT (Asymmetric)} & \textbf{0.6730±0.0056} \\
\midrule
\multirow{2}{*}{MalNet-Tiny} & \multirow{2}{*}{Sparse} 
& Symmetric  & 0.923±0.008 \\
& & \textbf{DARTS-GT (Asymmetric)} & \textbf{0.9334±0.0057} \\
\bottomrule
\end{tabular}%
}
\vspace{-2mm}
\end{table}


\subsection{Interpretability Analysis}
\label{subsection:Interpretability1}

Although performance gains demonstrate DARTS-GT's effectiveness, modern applications increasingly demand interpretable decisions. Having established DARTS-GT's competitive performance, in this section, we examine whether these heterogeneous architectures offer advantages in model transparency using the framework introduced in Section~\ref{method:interpret}.

\subsubsection{Quantitative Interpretability Metrics}

Table~\ref{tab:interpretability_metrics} presents the interpretability metrics derived from our causal ablation framework (Section~\ref{method:interpret}). Through systematic head masking on each test instance, we compute head deviations that quantify causal importance, then aggregate using median values for robustness to outliers. The median Specialization captures the typical concentration of head importance across instances, while median Focus measures consensus among important heads on critical nodes. Supplementary Figures 4 and 5 show population-level distributions as scatter plots, with median values indicated, for all tests under consideration.
 We evaluated on two diverse benchmarks: MolHIV (OGB, molecular graphs, classification) and Pep-Struc (LRGB, peptide graphs, regression), representing different domains, graph sizes, and task types. For transparency, we report GPS~\cite{rampa_recipe_nodate} with both sparse attention (matching our implementation) and dense attention (authors' original) on MolHIV, revealing how attention mechanism choices significantly impact interpretability ( Focus: 0.5734 sparse vs 0.111 dense).

    Higher Focus facilitates identifying critical nodes as important heads converge on common graph regions, while higher Specialization enables distinguishing which heads drive predictions. The combination of these metrics determines the practical effort required to understand model decisions.

DARTS-GT demonstrates consistent interpretability across both datasets, achieving best Specialization on MolHIV (0.00124) and best Focus on Pep-Struc (0.129), while securing second-best performance on the complementary metrics. This consistency across diverse domains suggests that DARTS-GT's produces reliably interpretable models. In contrast, GPS and Vanilla-GT show more variable patterns: GPS (sparse) achieves the best Focus but minimal Specialization on MolHIV, making head importance indistinguishable, while Vanilla-GT shows the best Specialization but lowest Focus on Pep-Struc, requiring reconciliation of divergent attention patterns.

These metrics characterize the complexity of the interpretation rather than establishing absolute superiority. Our evaluation on two datasets provides initial empirical evidence on real-world applications but, like any empirical study, cannot guarantee trends generalize across all benchmarks.  However, one of our primary contributions, the interpretability framework itself, enables practitioners to quantify the difficulty of interpretation. Instance-level analysis (Supplementary Figures 4 and 5) reveals considerable variation across graphs, reminding us that specific instances may present unique interpretability challenges regardless of model-level trends.

\begin{table}[!t]
\centering
\caption{Quantitative interpretability metrics across architectures. Median Specialization: concentration of importance in fewer heads; Median Focus: consensus on important nodes. Best values in \textbf{bold} while second-best are \underline{underlined}.}
\label{tab:interpretability_metrics}
\setlength{\tabcolsep}{4pt}
\begin{tabular}{l|l|ccc}
\toprule
\textbf{Dataset} & \textbf{Model} & \textbf{Specialization} & \textbf{Focus}  \\
\midrule
\multirow{3}{*}{MolHIV} 
    & Vanilla-GT (sparse) & 0.000634 & 0.533  \\
    & GPS (sparse) & 0.000044 & \textbf{0.5734}  \\
    & GPS (dense) & \underline{0.000091} & 0.111  \\
    & \textbf{DARTS-GT} (sparse) & \textbf{0.00124} & \underline{0.555} \\
\midrule
\multirow{3}{*}{Pep-Struc} 
    & Vanilla-GT (dense) & \textbf{0.055} & 0.067  \\
    & GPS (dense) & 0.038 & \underline{0.085} \\
    & \textbf{DARTS-GT} (dense) & \underline{0.044} & \textbf{0.129}  \\
\bottomrule
\end{tabular}
\end{table}

In our next section, we demonstrate how our framework enables granular, instance-specific interpretation grounded in causal head ablation, moving beyond aggregate statistics to understand individual prediction drivers.

\begin{figure*}[htbp]
\centering
\subfloat[DARTS-GT]{\includegraphics[width=0.75\textwidth]{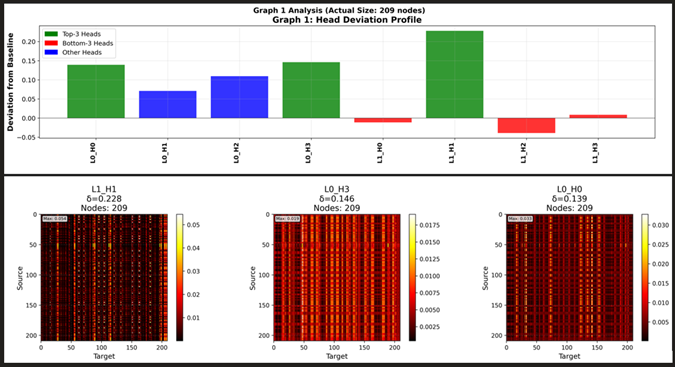}\label{fig:g1_darts}}\\
\subfloat[GPS]{\includegraphics[width=0.75\textwidth]{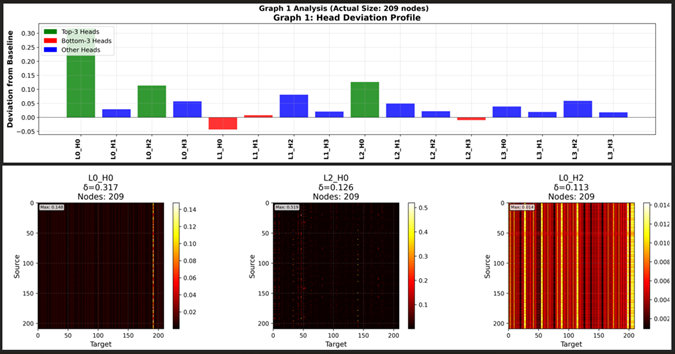}\label{fig:g1_gps}}\\
\subfloat[Vanilla-GT]{\includegraphics[width=0.75\textwidth]{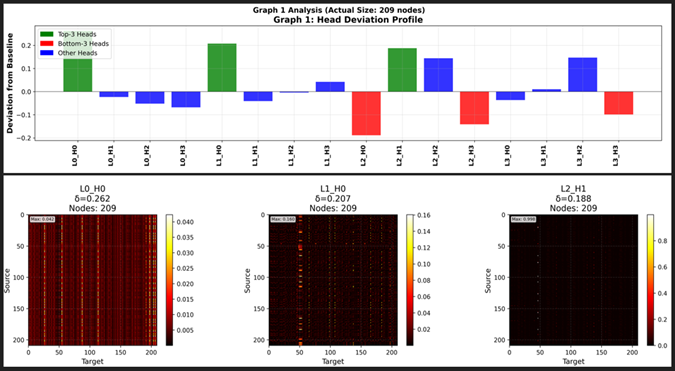}\label{fig:g1_vanilla}}
\caption{Instance analysis of Graph 1 (Peptides-Struc, 209 nodes) demonstrating deviation profiles and top-3 attention heatmaps.In the attention heatmaps, columns represent target nodes that receive attention. }
\label{fig:graph1_comparison}
\end{figure*}

\subsection{Instance-Specific Interpretation Analysis}
\label{subsection:Interpretability2}
Having established the interpretability characteristics at the dataset level, we now demonstrate how our framework provides actionable, instance-specific insights to understand individual predictions. Unlike aggregate metrics that characterize general model behavior, instance-level analysis reveals which specific graph components causally drive predictions for particular test cases through systematic head ablation. This approach is analogous to how GNNExplainer~\cite{ying_gnnexplainer_2019} identifies critical subgraphs for GNNs but is adapted for Graph Transformer attention mechanisms. It grounds interpretation in causal evidence rather than correlational attention patterns, revealing not just where models attend but which attention heads causally influence outcomes.

Our framework automatically generates per-instance JSON outputs containing head deviations, top-k important heads (we keep $k=4$ for MolHIV and $k=3$ for Peptides-struct for our experiments), their attended nodes, attention distribution statistics, and interpretability metrics (Specialization, Focus). The higher absolute deviation ($|\delta|$) indicates greater head significance, i.e. removing such heads causes larger prediction errors, providing causal evidence of their contribution.

Additionally, we computed the standard deviation of each head's attention distribution to characterize its focus pattern. Although not a central metric, this simple statistic provides useful context: a low standard deviation indicates that a head distributes attention uniformly across nodes (broad emphasis), while a high standard deviation suggests concentrated attention on specific nodes (selective emphasis). 

Figure \ref{fig:graph1_comparison} presents a representative instance from the Peptides-Struc datasets, where deviation profiles identify truly important heads regardless of visual prominence, and attention heatmaps of these top-k heads reveal critical graph regions. Each attention head is uniquely identified by its layer and head index, following the format \texttt{L<layer\_index>\_H<head\_index>}. For example, L0\_H0 refers to the first head in the first layer. While we demonstrate Peptides-struct example here, MolHIV is moved to supplementary (Supplementary Figure 3) for space constraints. 

Graph 1 (Figure~\ref{fig:graph1_comparison}, Peptides-Struct, 209 nodes) demonstrates the attention visualization paradox where visual prominence does not always correlate with importance. \textit{(a) DARTS-GT:} Top heads L1\_H1 ($\delta=0.228$), L0\_H3 ($\delta=0.146$), L0\_H0 ($\delta=0.139$) all exhibit low standard deviation of their attention distributions  (stdev) (0.0076, 0.0029, 0.0038) producing uniformly colorful attention patterns, while targeting different nodes  as L1\_H1 focuses most on nodes 27-28, L0\_H3 on node 47,197 and L0\_H0 on nodes 32,141. Moderate specialization (Spec$=0.085$) with low Focus ($0.059$) indicates distinguishable heads attending to disparate regions.
\textit{(b) GPS:} Demonstrates the paradox clearly—the most important head L0\_H0 ($\delta=0.317$, stdev $=0.0063$) appears dark/sparse concentrating most on node 191 (attention score 17.93) followed by 202 (attention score 2.20), while visually prominent colorful head with overall uniform attention, such as L0\_H2 ($\delta=0.113$, stdev $=0.0026$)  have minimal impact with a much weaker attention on node 197 (attention score 2.657). Very low Focus ($0.036$) further confirms disparate mechanisms.  
\textit{(c) Vanilla-GT: }Inverts the GPS pattern—highest impact head L0\_H0 ($\delta=0.262$, stdev $=0.0043$) shows uniform/colorful attention across nodes 54,203,26,113 and more (attn score around 4.0) while L2\_H1 ($\delta=0.188$, stdev $=0.0179$) displays a concentrated dark pattern on node 46 (attn score 13.07) despite lower importance. Highest specialization (Spec$=0.127$) with low Focus ($0.064$) suggests strongly differentiated but uncoordinated heads for this particular instance under this method.  

  To our knowledge, the above is the first instance-specific quantitative method for Graph Transformers to identify which heads and nodes causally drive predictions. We have quantitatively shown further (Supplementary Figure 1, Supplementary Section I), that entropy heuristics cannot replace our head-deviation based causal ablation method. Although space constraints limit us to two representative examples (additional instances available in our GitHub repository), these were strategically selected: Graph 1 (Peptides-struct) showcases a deliberate paradox where visual prominence do not always correlate with causal importance while Graph 50 (MolHIV) presented in supplementary section III, demonstrates our framework's standard operation and how these metrics varies based on attention types. 
  
Across instances, we did not observe a reliable correlation between attention salience and causal head importance. In GPS, the most critical head (L0\_H0, $\delta=0.317$) appears sparse/dark while colorful heads contribute minimally. In contrast, in Vanilla-GT, a visually prominent dark pattern (L2\_H1) proves causally less important than a uniform/colorful head (L0\_H0,  $\delta=0.262$). \textit{This instance-specific variability, where concentrated attention may be causally critical, irrelevant, or anywhere between, demonstrates that practitioners cannot determine head importance from visualization alone. The quantitative deviation analysis of our framework resolves this ambiguity by directly measuring the casual contribution of each head through ablation, regardless of visual appearance.} This enables reliable identification of functionally significant components that attention visualization would miss or misrepresent. Users can generate similar analyses for any instance using our code provided, extending beyond these illustrative examples to their specific use cases.

\section{Conclusion and Future Directions}

This work introduced DARTS-GT, a differentiable architecture search framework for Graph Transformers that discovers optimal depth-wise GNN configurations while providing quantifiable interpretability. We now revisit the research questions posed at the outset.

\textbf{Addressing Q1 - Architectural Performance and Adaptability:} Our experiments demonstrate that Graph Transformers can achieve superior performance through dynamic, layer-wise GNN selection. DARTS-GT consistently outperformed vanilla Graph Transformers on all benchmarks, with relative improvements ranging from 0.7\% to 318\% depending on the complexity of the dataset. Against state-of-the-art methods, DARTS-GT achieved the best new results on 4 of 8 datasets (ZINC, MalNet-Tiny, Peptides-Func, and Peptides-Struc), while remaining competitive on others. 

The architectures discovered reveal DARTS' adaptive nature: selection patterns range from highly specialized (81\% GINE on MolPCBA) to nearly uniform distributions (balanced usage on ZINC). This dataset-specific optimization validates that no universal architecture exists; rather, optimal configurations emerge from the interplay between task requirements and graph properties. Our ablation studies further confirmed that the asymmetric attention design (Q from features, K/V from GNNs) consistently outperforms symmetric variants, demonstrating that decoupling structural encoding from feature queries improves graph understanding. This suggests that without auxillary MPNNs and by redesigning the attention mechanism itself, comparable SOTA performance can be achieved.

\textbf{Addressing Q2 - Quantifiable Interpretability:} Our interpretability framework provides the first quantitative method to understand Graph Transformer decisions at both the population and the instance levels. The median-based aggregation of Specialization and Focus metrics offers robust dataset-level characterization. Practitioners can quickly assess whether the prediction of a model will be easily interpretable (high Specialization, high Focus) or require extensive analysis (low values on both). These population-level metrics serve as practical guidelines for model selection in deployment scenarios where interpretability matters.

At the instance level, our framework reveals a critical finding: visual attention salience does not reliably indicate functional importance. Through systematic ablation, we demonstrated cases where sparse, nearly invisible attention patterns drive predictions, while visually prominent patterns prove dispensable. This invalidates current visualization-based interpretation practices~\cite{el_towards_2025,hussain_global_2022} and establishes causal deviation analysis as essential to understand Graph Transformer decisions.

\textbf{Future Directions:} The convergence of improved performance and quantifiable interpretability opens exciting possibilities for scientific discovery. An appropriate real-world impactful domain application to test our DARTS-GT with the interpretability framework would be the protein structure-to-function prediction problem, where given a protein with known functions $[F1, F2, F3, F4]$, DARTS-GT could not only predict these functions accurately, but also identify which structural motifs causally contribute to each function. When the model makes a confident prediction on a novel protein, our interpretability framework can distinguish whether the prediction relies on established structural patterns (suggesting genuine functional similarity) or spurious correlations (indicating potential overfitting). 

In conclusion, DARTS-GT demonstrates that Graph Transformers do not need to choose between performance and interpretability. By redesigning the attention mechanism through differentiable architecture search and grounding interpretation in causal analysis, we provide a path toward models that are both more capable and more trustworthy: essential qualities for deployment in high-stakes scientific and medical applications.
\section{Acknowledgments}
This research was funded by Google DeepMind, with S.C. being supported by Google DeepMind Scholarship for her DPhil studies. The authors also thank Dr.Petar Veličković for helpful discussions. For the purpose of Open Access, the authors have applied a CC BY public copyright license to any Author Accepted Manuscript (AAM) version arising from this submission. The authors acknowledge the use of the University of Oxford Advanced Research Computing (ARC) facility in carrying out this work. \href{https://doi.org/10.5281/zenodo.22558}{https://doi.org/10.5281/zenodo.22558}.

\bibliographystyle{IEEEtran}
\bibliography{references}


\section*{Supplementary material}

# Supplementary Materials - DARTS-GT: Differentiable Architecture Search for Graph Transformers with Quantifiable Instance-Specific Interpretability Analysis

Shruti Sarika Chakraborty Peter Minary

## I. Relationship between Attention Concentration and Functional Importance

We investigate if heads with concentrated attention patterns exhibit higher functional importance. We test this systematically by correlating attention entropy with head deviation across all test instances for one representative dataset : Peptides-Struc for one method - DARTS-GT.

*a) Methodology.:* For each graph $G_i$ with $n$ nodes, head $m$ in layer $\ell$, we compute attention entropy from the attention matrix $A_m^{(\ell)} \in \mathbb{R}^{n \times n}$. First, we sum incoming attention for each target node:

$$S_m^{(\ell)}[j] = \sum_{k=1}^{n} A_m^{(\ell)}[k, j] \tag{1}$$

Then normalize to obtain a probability distribution:

$$P_m^{(\ell)}[j] = \frac{S_m^{(\ell)}[j]}{\sum_{j'=1}^{n} S_m^{(\ell)}[j']} \tag{2}$$

Finally, compute Shannon entropy:

$$H(m, \ell, i) = -\sum_{j=1}^{n} P_m^{(\ell)}[j] \log P_m^{(\ell)}[j] \tag{3}$$

where low entropy indicates concentrated attention on few nodes and high entropy indicates diffuse attention. For each instance $i$, we compute the Spearman rank correlation between $\{H(m, \ell, i)\}_{m \in \mathcal{M}, \ell \in \mathcal{L}}$ and $\{\delta_{m,\ell,i}\}_{m \in \mathcal{M}, \ell \in \mathcal{L}}$ across all heads and layers.

*b) Results on Peptides-Struct:* We apply this analysis to DARTS-GT on the Peptides-Struct dataset. Supplementary Figure 1 shows the distribution of per-instance correlations. The left histogram reveals correlations spanning from -1.0 to +1.0, with a median of only 0.167 and standard deviation of 0.437. Notably, the standard deviation exceeds twice the median value. The right boxplot confirms this inconsistency: the interquartile range crosses zero, and whiskers extend across nearly the full correlation range. Substantial portions of instances exhibit negative correlations (where concentrated attention corresponds to low importance), near-zero correlations (no relationship), and positive correlations (concentrated attention corresponds to high importance).

Department of Computer Science, University of Oxford, Oxford OX1 3QG, UK. Emails: shruti.chakraborty@cs.ox.ac.uk, peter.minary@cs.ox.ac.uk.

The weak median correlation (0.167) combined with high variance (0.437) demonstrates that attention concentration (entropy) is not a reliable predictor of functional importance. The relationship varies fundamentally across instances i.e. concentrated attention may indicate critical heads, dispensable heads, or have no relationship to importance at all. *This proves that attention-based visualization or entropy heuristics cannot substitute for causal ablation methods.* Our head deviation framework remains necessary for faithful interpretation, as it directly measures functional contribution rather than relying on correlational attention patterns that may be misleading.

## II. Ablation Studies: Dataset-Dependent Role of Edge Information

Here we examine the impact of edge residual connections across different attention mechanisms and datasets. While our GNN experts (GINE, GatedGCN, GATV2) inherently utilize edge features, these ablation studies demonstrate whether adding explicit edge residual connections to the attention mechanism provides additional benefits. Supplementary Table I presents results across all datasets comparing dense and sparse attention with and without edge residual connections. For computational efficiency, ZINC experiments use 1000 epochs, sufficient to demonstrate relative performance trends. The results reveal dataset-specific utility of edge residuals. These findings demonstrate that edge residual connections are not universally beneficial, validating our design choice of making them optional.

## III. Instance-Specific Interpretability Analysis for MolHIV :

Here we demonstrate instance-specific interpretability analysis for MolHIV's graph number: 50. For Graph-50 (17 nodes) of MolHIV, head deviation profiles and attention patterns (Supplementary Figure 3) reveal distinct interpretability characteristics. In the attention heatmaps, columns represent target nodes that receive attention. *(a) DARTS-GT:* Clear head specialization (Spec= 0.00288) with top heads forming two consensus pairs - L0_H3 ($\delta = 0.016$ - highest) /L1_H1 attending to target node 1, L2_H0/L1_H0 attending to target node 9. The Focus for this instance is 0.333. *(b) GPS with dense attention:* Minimal head specialization (Spec= $1.01 \times 10^{-9}$) with nearly imperceptible deviations (highest $\delta = 4.87 \times 10^{-9}$ at L5_H1),

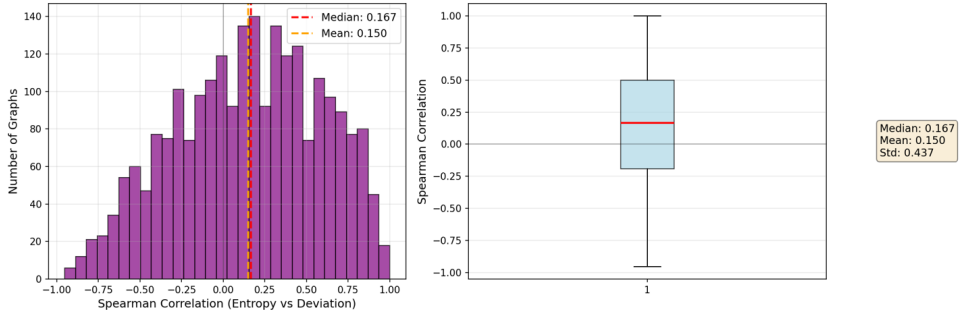

Supplementary Figure 1: Distribution of per-instance Spearman correlations between attention entropy and head deviation on Peptides-Struct.

Supplementary Table I: Effect of edge features on Graph Transformer performance across multiple datasets and attention types. Shown is the mean ± std across 3 seeds. Best results are in **bold**.

| Dataset | Attn Type | Mode | Perf |
|---------|-----------|------|------|
| ZINC* | Dense | No edge | 0.1054±0.0055 |
| | | Edge | **0.098±0.0082** |
| | Sparse | No edge | 0.0812±0.0043 |
| | | Edge | **0.0782±0.006** |
| MOLPCBA | Dense | No edge | 0.2011±0.0101 |
| | | Edge | **0.2224±0.004** |
| | Sparse | No edge | **0.258±0.0043** |
| | | Edge | 0.2545±0.003 |
| CLUSTER | Dense | No edge | 0.2693±0.0013 |
| | | Edge | **0.3107±0.0122** |
| | Sparse | No edge | 0.7773±0.0020 |
| | | Edge | **0.7835±0.0004** |
| Peptides-Struc | Dense | No edge | **0.2466±0.0003** |
| | | Edge | 0.248±0.0025 |
| | Sparse | No edge | 0.263±0.0004 |
| | | Edge | 0.263±0.0008 |

suggests uniform head importance. Partial consensus with L5_H1/L0_H1 attending to target node 10 while others target nodes 7,11. The overall Focus attained is 0.167. *(c) GPS with sparse attention:* Low head specialization (Spec= $7.39 \times 10^{-5}$) with low deviations (highest $\delta = 0.0003$ at L4_H3) . High consensus with L4_H3/L6_H0/L5_H3 attending to target node 9 while L6_H2 target node 14. The overall Focus attained is very high 0.5. *(d) Vanilla-GT:* Low specialization (Spec= $2.47 \times 10^{-5}$). Similar partial consensus with L14_H1/L3_H2 attending to target node 14, others targeting nodes 8,5 and overall Focus is same as GPS (dense attention) with 0.167. Despite similar Focus values in GPS (dense attn) and Vanilla-GT, their vastly different Specialization values indicate distinct interpretation challenges.

## IV. LAYER-WISE ARCHITECTURE DISCOVERY VIA DARTS

The DARTS search process automatically discovers optimal GNN operators for each layer, as illustrated in Supplementary Figure 2 for CLUSTER and ZINC datasets. These results demonstrate that different layers require different inductive

biases, with the search adapting operator selection based on both network depth and dataset characteristics.

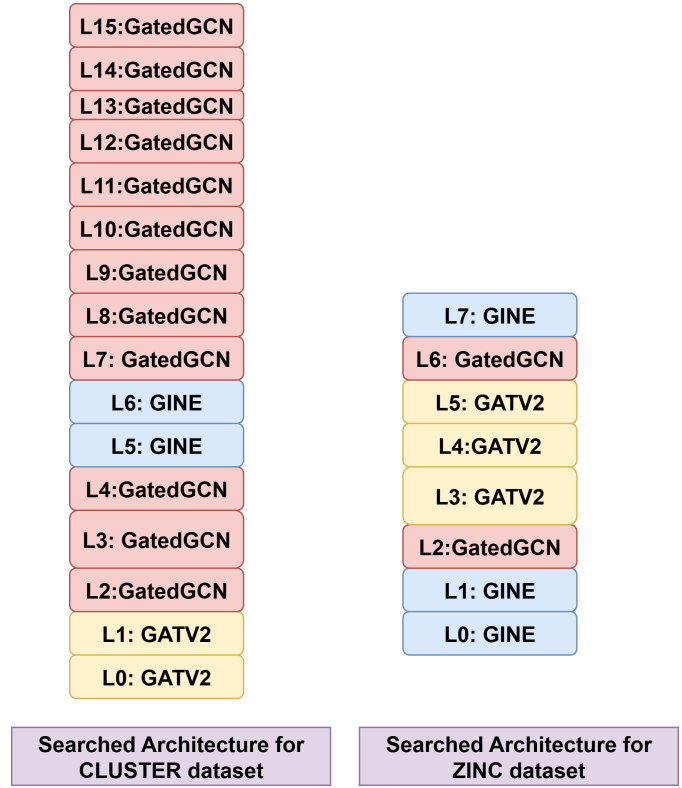

Supplementary Figure 2: **Discovered layer-wise operators after DARTS.** For each layer, the search selects exactly one operator from {GINE, GATv2, GatedGCN}. *CLUSTER (left, L0–L15):* the architecture is dominated by **GatedGCN** across depth, with brief **GINE** selections around L5–L6 and **GATv2** at L0–L1. *ZINC (right, L0–L7):* shallow layers favor **GINE** (L0–L1, L7), the mid-depth region prefers **GATv2** (L3–L5), and **GatedGCN** appears at L2 and L6. These dataset-specific, depth-varying patterns indicate that the search adapts inductive biases per layer, rather than converging to a uniform choice. After search, the selected per-layer GNN is used once to produce $K, V$ and is shared across heads.

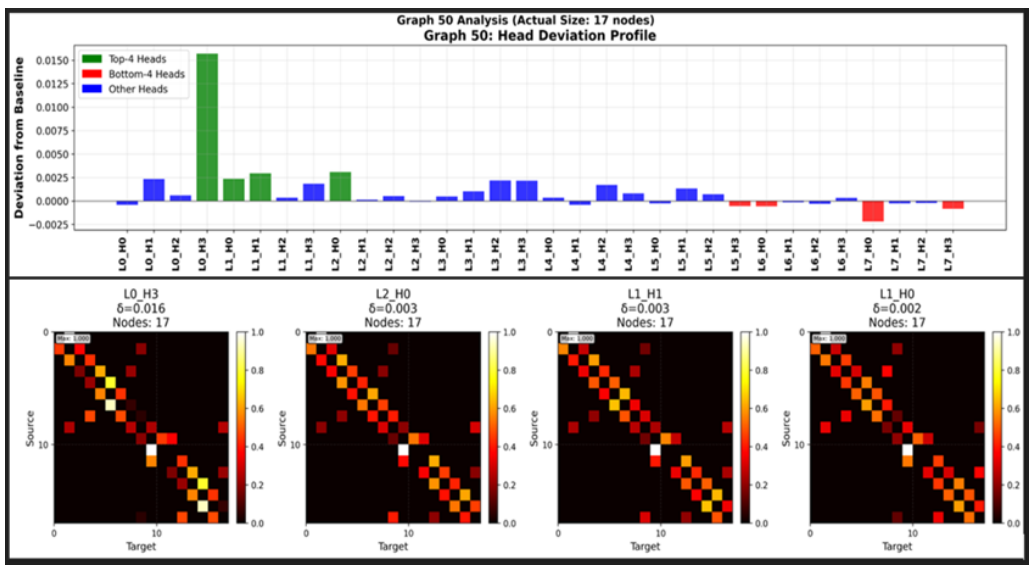

(a) DARTS-GT

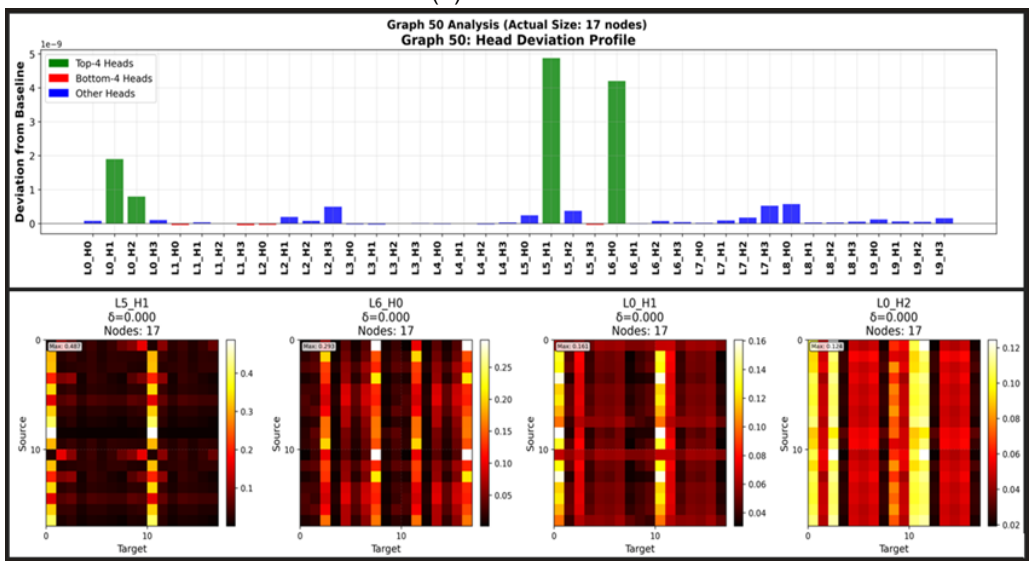

(b) GPS

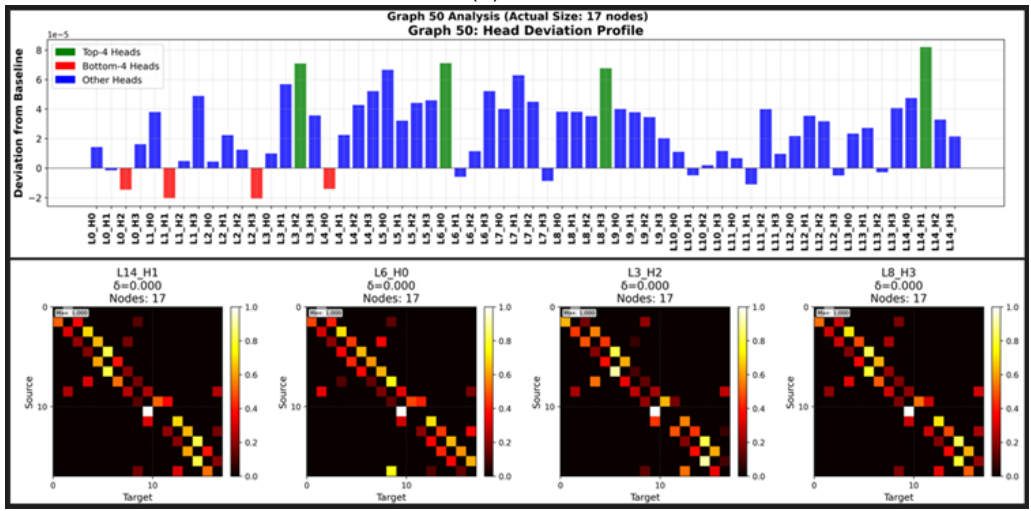

(c) Vanilla-GT

Supplementary Figure 3: Instance analysis of Graph 50 (MolHIV, 17 nodes). Deviation profiles and top-4 attention heatmaps for each model, with columns representing target nodes. GPS with dense attention is given here.

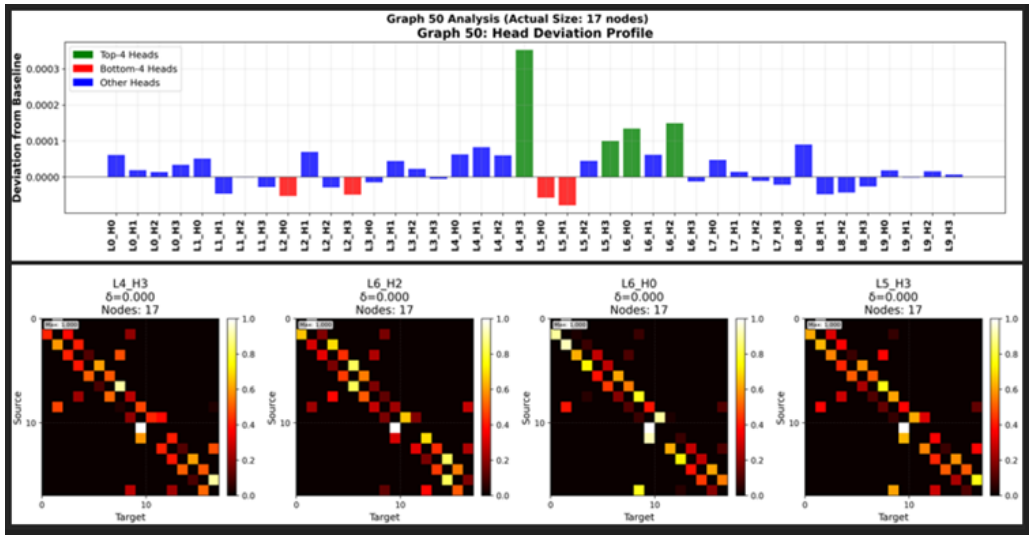

(d) GPS (sparse)

Supplementary Figure 3: (continued) Instance analysis of Graph 50 - GPS with sparse attention.

## V. POPULATION LEVEL INTERPRETABILITY ANALYSIS

Supplementary figures 4 and 5 visualize population-level interpretability distributions across two test sets : Peptide-Struc and MolHIV respectively, revealing distinct patterns between architectures and datasets. The plots demonstrate that interpretability is inherently dataset-dependent and varies significantly at the instance level. While we report median values for dataset-level comparison, each point represents an individual graph with its own Specialization and Focus values derived from our causal ablation framework. This instance-level variation, visible as the scatter around median values, underscores that some graphs remain challenging to interpret regardless of model choice, while others yield clear interpretability patterns.

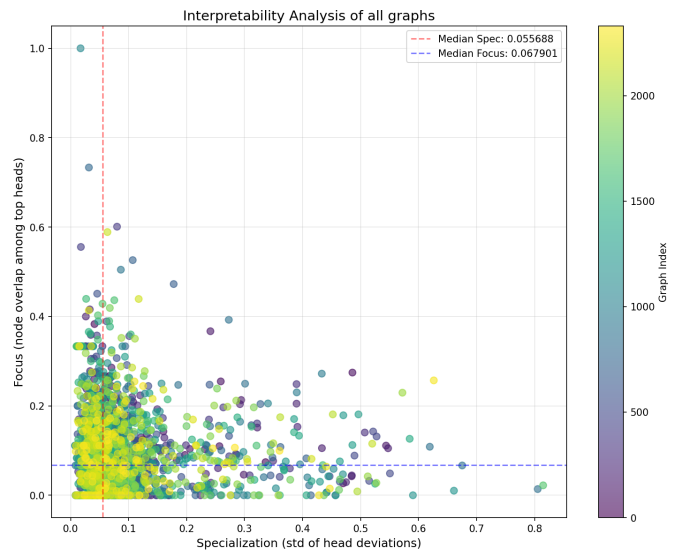

Supplementary Figure 4: Instance-level interpretability distributions on Peptides-Struc test set. (a) Vanilla-GT model (dense attention). Each point represents a single test graph positioned by its Specialization (head importance concentration) and Focus (consensus among important heads).

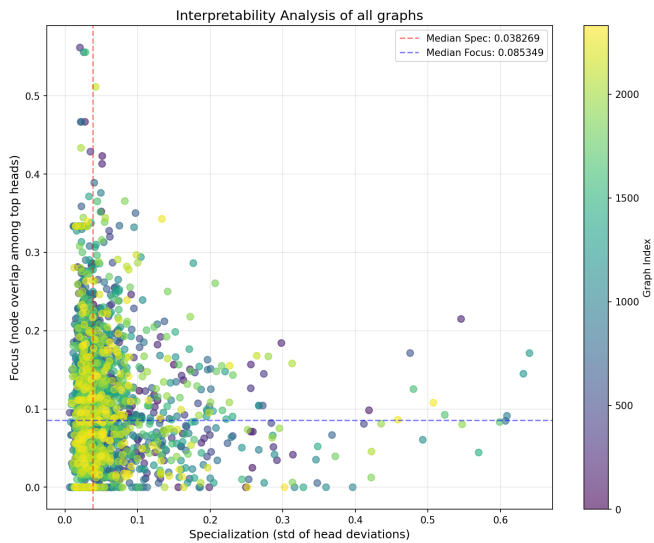
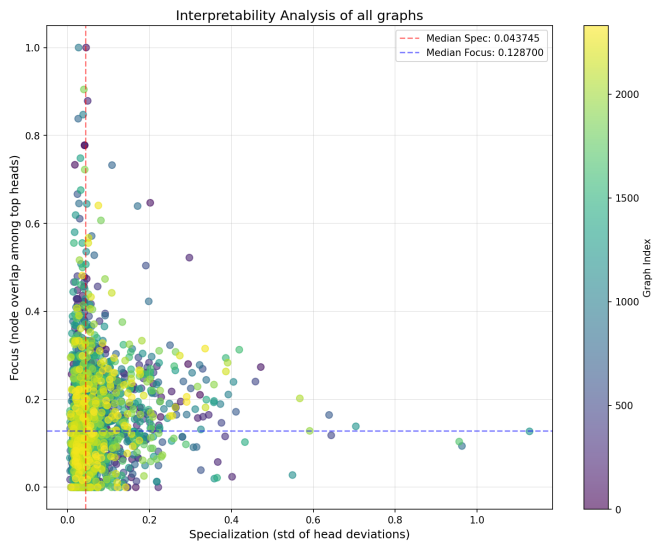

(a) GPS (Pep-Struc, dense)                              (b) DARTS-GT (Pep-Struc, dense)

Supplementary Figure 4: (continued) (b) GPS model shows different clustering patterns compared to Vanilla-GT. (c) DARTS-GT achieves the highest median Focus (0.129), indicating stronger node consensus. All models use dense attention. Note the different scale compared to MolHIV (Supplementary Figure 5), with Specialization value, suggesting dataset-specific interpretability characteristics.

Supplementary Table II: Hyperparameters across all datasets (Part 1)

| Hyperparameters | ZINC | MolHIV | MOLPCBA | MALNET |
|---|---|---|---|---|
| *Architecture Parameters* | | | | |
| Number of Layers | 8 | 8 | 4 | 4 |
| Hidden dim | 64 | 64 | 384 | 64 |
| Attention type | Sparse | Sparse | Sparse | Sparse |
| Heads | 4 | 4 | 4 | 4 |
| Dropout | 0 | 0.05 | 0.2 | 0 |
| Attention dropout | 0.5 | 0.5 | 0.5 | 0.5 |
| Edge residual weight | 0.1 | NA | NA | 0.1 |
| Q-projection | Identity | Identity | Identity | Identity |
| *DARTS Parameters* | | | | |
| DARTS epochs | 120 | 30 | 30 | 30 |
| DARTS splits ratio | 0.6 | 0.6 | 0.6 | 0.6 |
| Architecture LR | 4.00E-04 | 4.00E-04 | 4.00E-04 | 4.00E-04 |
| Grad clip | 5 | 5 | 5 | 5 |
| *Training Parameters* | | | | |
| LR reduce factor | 0.5 | 0.5 | 0.5 | 0.5 |
| LR schedule patience | 10 | 10 | 10 | 10 |
| Min learning rate | 1.00E-06 | 1.00E-06 | 1.00E-06 | 1.00E-06 |
| Initial learning rate | 0.0025 | 0.0025 | 0.0025 | 0.0025 |
| Weight decay | 3.00E-04 | 3.00E-04 | 3.00E-04 | 3.00E-04 |
| Epochs | 2000 | 100 | 100 | 100 |
| Warmup epochs | 50 | 5 | 5 | 5 |
| *Graph-specific Parameters* | | | | |
| Graph pooling | sum | mean | mean | max |
| Positional Encoding | RWSE-20 | RWSE-16 | RWSE-16 | None |
| PE dim | 28 | 16 | 20 | NA |
| PE encoder | linear | linear | linear | NA |
| *Computational Resources* | | | | |
| DARTS-Parameters | 641989 | 583017 | 10186472 | 327893 |
| Discrete-Parameters | 412397 | 344929 | 5707484 | 208329 |
| Total DARTS time | 4490.0s | 2777.0s | 4441.0s | 630.0s |
| Discrete phase time | 26.0s/15.0h | 63.0s/2.37h | 158.0s/5.0h | 16.0s/0.645h |

Supplementary Table III: Hyperparameters across all datasets (Part 2)

| Hyperparameters | PATTERN | CLUSTER | PEP-FUNC | PEP-STRUC |
|---|---|---|---|---|
| *Architecture Parameters* | | | | |
| Number of Layers | 6 | 16 | 2 | 2 |
| Hidden dim | 64 | 48 | 120 | 120 |
| Attention type | Sparse | Sparse | Dense | Dense |
| Heads | 4 | 8 | 4 | 4 |
| Dropout | 0 | 0 | 0 | 0 |
| Attention dropout | 0.5 | 0.5 | 0.5 | 0.5 |
| Edge residual weight | 0.1 | 0.1 | NA | NA |
| Q-projection | Identity | Identity | MLP | MLP |
| *DARTS Parameters* | | | | |
| DARTS epochs | 20 | 25 | 50 | 50 |
| DARTS splits ratio | 0.6 | 0.6 | 0.6 | 0.6 |
| Architecture LR | 4.00E-04 | 4.00E-04 | 4.00E-04 | 4.00E-04 |
| Grad clip | 5 | 5 | 5 | 5 |
| *Training Parameters* | | | | |
| LR reduce factor | 0.5 | 0.5 | 0.5 | 0.5 |
| LR schedule patience | 10 | 10 | 10 | 10 |
| Min learning rate | 1.00E-06 | 1.00E-06 | 1.00E-06 | 1.00E-06 |
| Initial learning rate | 0.0025 | 0.0025 | 0.0025 | 0.0025 |
| Weight decay | 3.00E-04 | 3.00E-04 | 3.00E-04 | 3.00E-04 |
| Epochs | 100 | 100 | 200 | 200 |
| Warmup epochs | 5 | 5 | 0 | 0 |
| *Graph-specific Parameters* | | | | |
| Graph pooling | NA | NA | mean | mean |
| Positional Encoding | LapPE-16 | LapPE-10 | LapPE-10 | LapPE-10 |
| PE dim | 16 | 16 | 16 | 16 |
| PE encoder | DeepSet | DeepSet | DeepSet | DeepSet |
| *Computational Resources* | | | | |
| DARTS-Parameters | 487369 | 728678 | 604817 | 604696 |
| Discrete-Parameters | 338775 | 511250 | 426341 | 448240 |
| Total DARTS time | 1320.45s | 3454.26s | 409.0s | 411.54s |
| Discrete phase time | 58.0s/1.87h | 97.4s/3.88h | 12.49s/0.775h | 12.5s/0.845h |

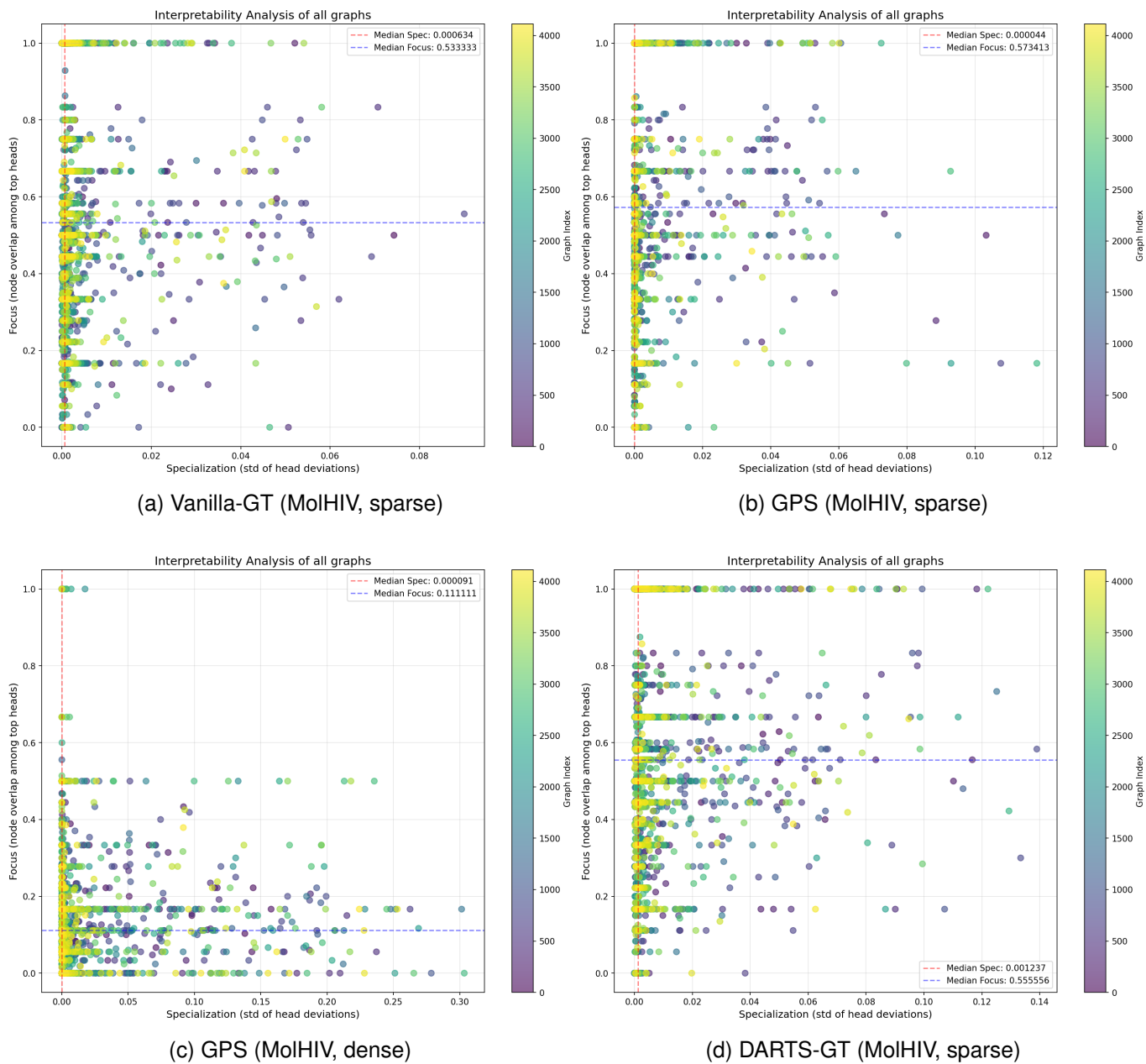

(a) Vanilla-GT (MolHIV, sparse)   (b) GPS (MolHIV, sparse)

(c) GPS (MolHIV, dense)   (d) DARTS-GT (MolHIV, sparse)

Supplementary Figure 5: Instance-level interpretability distributions on MolHIV test set. Each point represents a single test graph positioned by its Specialization (head importance concentration) and Focus (consensus among important heads). GPS shown with both sparse and dense attention to demonstrate architectural impact on interpretability.

Supplementary Table IV: Comprehensive parameter comparison across all models and datasets. Best (lowest) parameter count is highlighted in **bold**

| Model | Attention | ZINC | MolHIV | MOLPCBA | MALNET | PATTERN | CLUSTER | PEP-STRUC | PEP-FUNC |
|-------|-----------|------|--------|---------|--------|---------|---------|-----------|----------|
| *Dense Attention* | | | | | | | | | |
| Vanilla-GT | Dense | **3,56,261** | 4,51,793 | **84,06,236** | 5,03,749 | **3,02,385** | **3,85,126** | 4,29,611 | 4,29,490 |
| GPS | Dense | 4,23,717 | 5,58,625 | 97,44,496 | 5,27,237 | 3,37,201 | 5,02,054 | 5,04,459 | 5,04,362 |
| DARTS-GT | Dense | 4,41,751 | **3,98,289** | 85,57,152 | NA | 3,61,591 | 5,07,206 | **4,26,341** | 4,48,240 |
| *Sparse Attention* | | | | | | | | | |
| Vanilla-GT | Sparse | **3,56,261** | 4,51,793 | 84,06,236 | NA | **3,02,385** | 3,38,086 | 4,12,397 | 4,29,490 |
| DARTS-GT | Sparse | 4,28,333 | **3,44,929** | **57,07,484** | 2,08,329 | 3,38,775 | 5,11,250 | **3,97,301** | **4,08,160** |



\end{document}